\pdfoutput=1
\documentclass[11pt]{article}

\usepackage{authblk}
\usepackage[final]{acl}

\usepackage{times}
\usepackage{latexsym}

\usepackage[T1]{fontenc}
\usepackage{microtype}

\usepackage{inconsolata}

\usepackage[T1]{fontenc}    % use 8-bit T1 fonts
\usepackage{hyperref}       % hyperlinks
\usepackage{url}            % simple URL typesetting
\usepackage{booktabs}       % professional-quality tables
\usepackage{amsfonts}       % blackboard math symboonline.ac}       % compact symbols for 1/2, etc.
\usepackage{microtype}      % microtypography
\usepackage{color}
\usepackage{algorithm,algpseudocode}
\usepackage{caption}
\usepackage{mwe}
\usepackage{lipsum}

\usepackage{algorithm}
\usepackage{algpseudocode}
\usepackage{graphicx}
\usepackage{amsmath}
\usepackage{amssymb}
\usepackage{float}
\usepackage{stfloats}
\usepackage{bbm}
\usepackage{dsfont}
\usepackage{algorithm}
\usepackage{array}
\usepackage[caption=false,font=normalsize,labelfont=sf,textfont=sf]{subfig}
\usepackage{textcomp}

\usepackage{verbatim}
\usepackage{multirow}
\usepackage{graphicx}
\usepackage{pifont}
\newcommand{\cmark}{\ding{51}}%
\newcommand{\xmark}{\ding{55}}%
\usepackage{color}
\usepackage[hang,flushmargin]{footmisc} 
\usepackage{booktabs}
\usepackage[normalem]{ulem}

\usepackage{soul}
\usepackage{float}
\usepackage{tcolorbox}
\usepackage{tikz}

\soulregister{\cite}7
\soulregister{\ref}7
\soulregister\eqref7

\usepackage{colortbl}

\definecolor{azure(web)(azuremist)}{rgb}{0.94, 1.0, 1.0}
\definecolor{oldlace}{rgb}{0.99, 0.96, 0.9}
\definecolor{pearl}{rgb}{0.94, 0.92, 0.84}
\definecolor{seashell}{rgb}{1.0, 0.96, 0.93}
\definecolor{silver}{rgb}{0.75, 0.75, 0.75}
\definecolor{platinum}{rgb}{0.9, 0.89, 0.89}
\definecolor{almond}{rgb}{0.94, 0.87, 0.8}
\definecolor{lightskyblue}{RGB}{173, 216, 230}

\newcommand{\circledone}{\ding{192}}%
\newcommand{\circledtwo}{\ding{193}}%
\newcommand{\circledthree}{\ding{194}}%
\newcommand{\circledfour}{\ding{195}}%

\usepackage{arydshln}

\title{\textsc{UAlign}: Leveraging Uncertainty Estimations for Factuality Alignment on Large Language Models}

\author[\ding{171}]{\bf Boyang Xue}
\author[\ding{168}]{\bf Fei Mi}
\author[\ding{168}]{\bf Qi Zhu}
\author[\ding{171},\thanks{~Corresponding author.}]{\bf Hongru Wang}
\author[\ding{171}]{\bf Rui Wang}
\author[\ding{117}]{\bf Sheng Wang}
\author[\ding{115}]{\\ \bf Erxin Yu}
\author[\ding{118}]{\bf Xuming Hu}
\author[\ding{171},\ding{170}]{\bf Kam-Fai Wong}

\affil[\ding{171}]{The Chinese University of Hong Kong, $^\text{\ding{168}}$Huawei Noah’s Ark Lab}
\affil[\ding{117}]{The University of Hong Kong, $^\text{\ding{115}}$The Hong Kong Polytechnic University}
\affil[\ding{118}]{Hong Kong University of Science and Technology, Guangzhou}
\affil[\ding{170}]{MoE Key Laboratory of High Confidence Software Technologies
\authorcr \texttt{\{byxue, hrwang, kfwong\}@se.cuhk.edu.hk}}

\begin{document}
\maketitle
\begin{abstract}

Despite demonstrating impressive capabilities, Large Language Models (LLMs) still often struggle to accurately express the factual knowledge they possess, especially in cases where the LLMs' knowledge boundaries are ambiguous.
To improve LLMs' factual expressions, we propose the \textsc{UAlign} framework, which leverages \textbf{U}ncertainty estimations to represent knowledge boundaries, and then explicitly incorporates these representations as input features into prompts for LLMs to \textbf{Align} with factual knowledge.
First, we prepare the dataset on knowledge question-answering (QA) samples by calculating two uncertainty estimations, including confidence score and semantic entropy, to represent the knowledge boundaries for LLMs. 
Subsequently, using the prepared dataset, we train a reward model that incorporates uncertainty estimations and then employ the Proximal Policy Optimization (PPO) algorithm for factuality alignment on LLMs.
Experimental results indicate that, by integrating uncertainty representations in LLM alignment, the proposed \textsc{UAlign} can significantly enhance the LLMs' capacities to confidently answer known questions and refuse unknown questions on both in-domain and out-of-domain tasks, showing reliability improvements and good generalizability over various prompt- and training-based baselines.

\end{abstract}

\section{Introduction}
\label{sec:intro}

Despite the remarkable proficiency of large language models (LLMs) across a diverse range of tasks \citep{touvron2023llama,openai2023gpt4,vicuna2023},
they still frequently face challenges in accurately expressing factual knowledge that they learned from the pre-training stage but are uncertain about.
In such cases, the knowledge boundaries are somewhat ambiguous by LLMs, remaining a gap between ``known'' and ``expression'' \citep{lin2024flame,zhang2024selfalignment,li2024surveyhonestylargelanguage}, which may lead to the hallucination problem and undermine the reliability and applicability to users.

\begin{figure}[t!]
  \centering
  \includegraphics[width=0.40\textwidth]{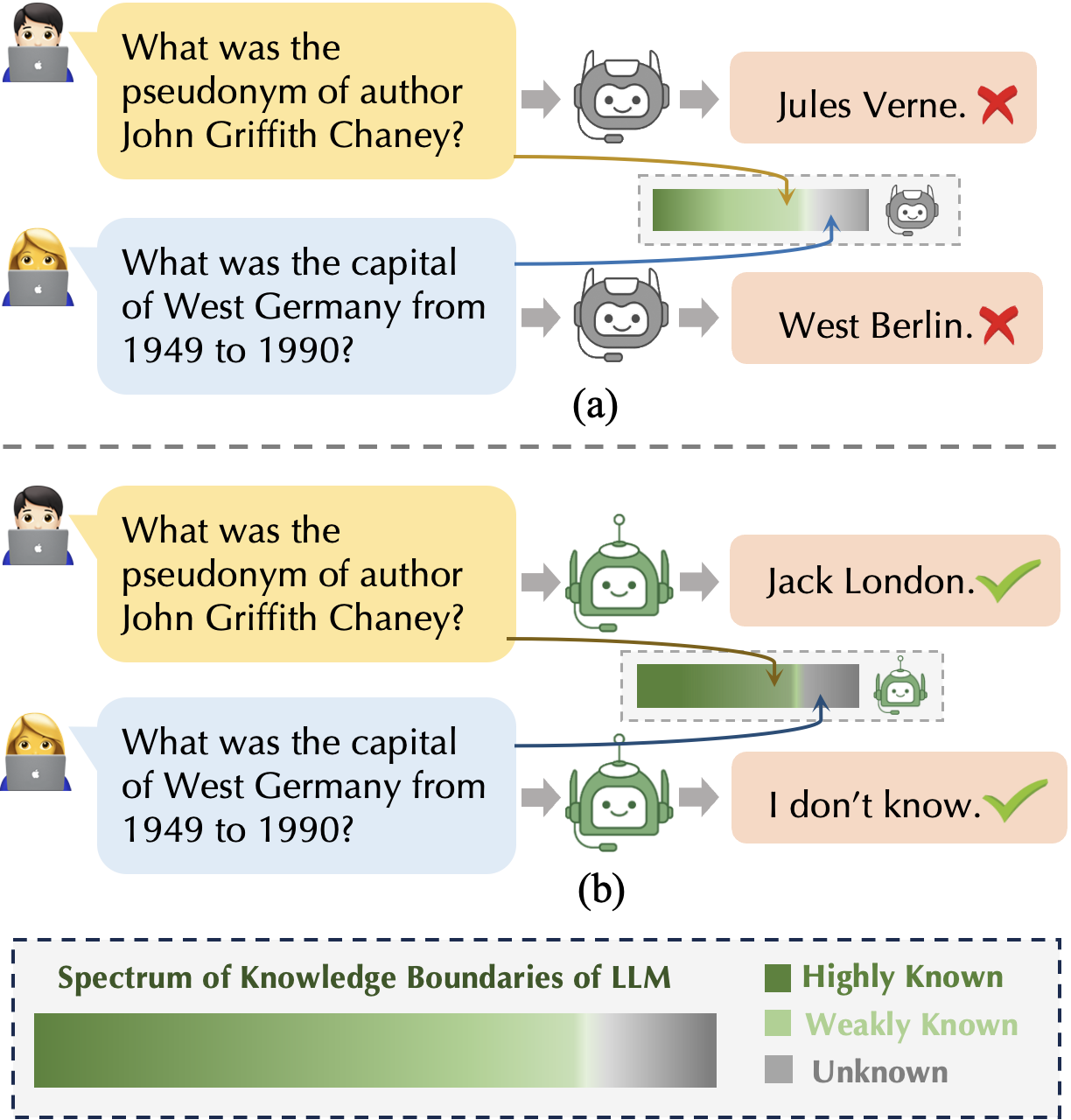}
    \caption{Examples of LLMs with (a) ambiguous and (b) explicit knowledge boundaries to answer questions.}
  \label{fig:frame}
\end{figure}

LLMs typically generate responses (``expression'') based on knowledge distributions learned during pre-training (``known'').
However, much of the knowledge acquired during this phase exhibits vague boundaries, comprising numerous learned but uncertain knowledge pieces (\textit{weakly known, light green area of spectrum} in Fig. \ref{fig:frame} (a)) \citep{gekhman-etal-2024-fine}.
% preventing LLMs from confidently conveying accurate information in downstream tasks when they hold relevant knowledge but don't make sure \citep{zhang2024selfalignment}.
Hence, LLMs may not confidently convey accurate information in downstream tasks even though they hold relevant knowledge but don't make sure \citep{zhang2024selfalignment}.
Additionally, LLMs may exhibit overconfidence in the knowledge they are unfamiliar with (\textit{unknown, the gray area of spectrum} in Fig. \ref{fig:frame} (a)), leading to fabricated or hallucinatory content \citep{zhang2024rtuning,liu2024examining}. 
This issue primarily arises from that LLMs don't properly reconcile the knowledge boundaries with factual accuracy during alignment \citep{tian2024finetuning}.
Unlike previous works that focused on reinforcement learning (RL) through knowledge feedback or factuality alignment \citep{liang-etal-2024-learning,xu2024rejection,tian2024finetuning,lin2024flame,zhang2024selfalignment,yang2024alignmenthonesty}, our objective is to elicit LLMs' weakly known facts and extend beyond merely discerning unknown facts by explicitly utilizing knowledge boundaries in alignment.
We aim to leverage the knowledge boundary information of LLMs to instruct LLMs to confidently express their known yet uncertain information and firmly refuse questions beyond their knowledge as in Fig. \ref{fig:frame} (b).
% thereby minimizing the discrepancy between ``known'' and ``expression''.
Based on improvements of ``known'', LLMs' expressions are more truthful and reliable, thereby minimizing the discrepancy between ``known'' and ``expression'' \citep{lin2024flame,zhang2024selfalignment,li2024surveyhonestylargelanguage}.

Inspired by the aforementioned analysis, we propose the \textsc{UAlign} framework, which strategically models \textbf{U}ncertainty regarding knowledge boundary representations, subsequently \textbf{Align}ing these estimations with factuality.
Therefore, the \textsc{UAlign} framework focuses on two pivotal issues: how to capture the knowledge boundary representations and how to align with factuality.

% \begin{figure*}[t!]
%   \centering
%   \includegraphics[width=0.99\textwidth]{figure/frame.png}
%     \caption{Examples of answering a question on the vanilla LLM (upper) and LLM using \textsc{UAlign} method.}
%   \label{fig:frame}
% \end{figure*}

First, we prepare the dataset that incorporates knowledge boundary information for alignment in the \textsc{UAlign} framework.
Knowledge boundaries always indicate the known level of factual knowledge, generally implemented using uncertainty estimation methods on LLMs \citep{ren2023investigating}.
To precisely capture the intrinsic perception of knowledge boundary representations given the knowledge QA datasets, we adopt two uncertainty estimations of accuracy-based confidence score \citep{xiong2024can} and semantic entropy \citep{kuhn2023semantic} respectively.
% We multiply sample responses to a question using varied prompting and temperature sampling strategies to alleviate overconfidence and approximate actual knowledge distribution to calculate confidence and entropy to each question.
We sample multiple responses to a question using varied prompting and temperature sampling to approximate actual knowledge boundaries by calculating the confidence and entropy of each question.
The two measures \citep{kuhn2023semantic,xiong2024can}, as complementary, can reflect the convince and dispersion of generated responses to a question based on LLMs' internal knowledge.
Questions with at least one correct sampled answer are regarded as ``known'', and those with all incorrect sampled responses are considered ``unknown''.
We revise ground-truth answers to unknown questions to refusal responses to delineate known and unknown facts \citep{zhang2024rtuning}.

Second, following \citet{ouyang2022training}, we explicitly leverage the uncertainty estimations to align with factuality on the prepared dataset using both supervised fine-tuning (SFT) and reinforcement learning (RL).
We employ SFT to train two uncertainty estimation models to predict confidence and entropy, and then train a reward model to evaluate the correctness of the generated answer conditioned on the input comprising the question, the generated response, and two uncertainty estimations regarding the knowledge boundary.
% With the reward model, we elicit LLMs' factual expressions by feeding both questions and two measures to LLMs using the Proximal Policy Optimization (PPO) \citep{schulman2017proximal} algorithm, as illustrated in Fig. \ref{fig:frame}.
With the reward model, we further adopt the Proximal Policy Optimization (PPO) \citep{schulman2017proximal} algorithm for LLM alignment by feeding both questions and two measures as prompts to elicit the policy LLM's factual expressions to improve the reliability.
% by feeding both questions and two measures to LLMs using the Proximal Policy Optimization (PPO) \citep{schulman2017proximal} algorithm.

Experiments are conducted to evaluate in-domain and out-of-domain performance on a range of knowledge QA datasets.
The results demonstrate our proposed \textsc{UAlign} method significantly enhances the reliability and generalization for LLMs over several baseline methods to accurately express known factual knowledge and refuse unknown questions, suggesting that leveraging the two employed uncertainty estimations in alignment can notably improve LLMs' factuality.
% even where the initial knowledge boundaries are inherently ambiguous.
% Improvements over several baseline methods indicate that \textsc{CUAlign} can better guide LLMs to align with verifiable factual knowledge by leveraging the certainty and uncertainty estimations.

In summary, our contributions are as follows.

1) To the best of our knowledge, \textsc{UAlign} is the first to explicitly leverage the uncertainty estimations representing knowledge boundaries for LLM alignment, heralding a promising direction for future research of LLM training
% \footnote{The codes will be released on GitHub.}.
\footnote{Codes are released on \href{https://github.com/AmourWaltz/UAlign}{https://github.com/AmourWaltz/UAlign}.}.

% 2) We introduce the UAlign dataset with a comprehensive dataset construction process to capture certainty and uncertainty estimations regarding knowledge boundaries to questions on LLMs.

2) We demonstrate that jointly incorporating confidence and semantic entropy into prompts can provide precise knowledge boundary information to elicit LLMs' factual expressions.

3) We conduct main experiments by comparing our \textsc{UAlign} with various baselines as well as ablation studies, validating the reliability improvements and robust generalization of the \textsc{UAlign} method.

\section{Methodology}
\label{sec:method}

\begin{figure*}[ht]
  \centering
  \includegraphics[width=.95\textwidth]{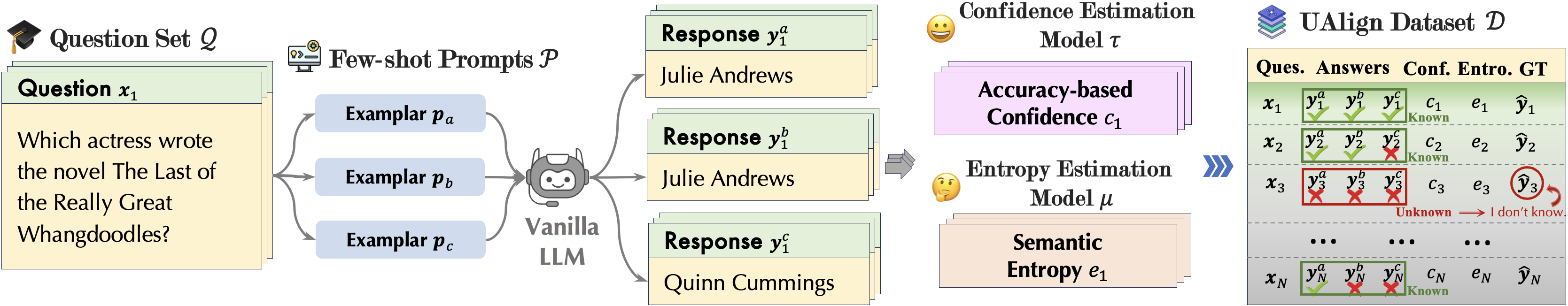}
    \caption{Illustration of \textsc{UAlign} dataset preparation process.}
  \label{fig:dataset}
\end{figure*}

The proposed \textsc{UAlign} framework is introduced in this section with two parts:
The Sec. \ref{ssec:data} involves the \textsc{UAlign} dataset preparation process, including strategies to collect multiple responses, as well as uncertainty measures to capture intrinsic representations of knowledge boundary on knowledge-based QA pairs as illustrated in Fig. \ref{fig:dataset}.
The Sec. \ref{ssec:align} utilizes the obtained \textsc{UAlign} dataset to train the uncertainty estimation models, and further explicitly incorporate the estimations as input features to elicit LLMs to generate factual responses using SFT- and PPO-based alignment methods as shown in Fig. \ref{fig:align} and Algorithm \ref{algorithm}.

\subsection{Dataset Preparation}
\label{ssec:data}

\subsubsection{Responses Sampling Strategy}
\label{sssec:response}

As in Fig. \ref{fig:dataset}, to explore the knowledge boundary of the LLM given a question, we sample multiple responses by repeating the generation procedure several times.
In this phase, the preparation process can be represented in a tuple $\left ( \mathcal{Q}, \mathcal{P}, \mathcal{A} \right )$.
$\mathcal{Q}$ contains a batch of $N$ QA pairs 
% $\left \{(\boldsymbol{x}_i, \boldsymbol{\hat y}_i)|i \in \left \{1, \dots, N \right \} \right \}$
${\left \{(\boldsymbol{x}_i, \boldsymbol{\hat y}_i) \right \}}_{i=1}^{N}$ 
where $\boldsymbol x_i$ and $\boldsymbol{\hat y}_i$ denote the $i$-th question and ground-truth answer respectively.
To mitigate context sensitivity, we utilize different few-shot prompts in $\mathcal{P}$ with temperature $T=0.2$ to make a trade-off between the accuracy and diversity to represent knowledge boundaries \citep{gekhman-etal-2024-fine}.
The few-shot prompt set $\mathcal{P}$ consists of $K$ different 1-shot exemplars in this work which is enough for LLMs to generate answers in the correct format.
We present the few-shot prompts for sampling on TriviaQA and SciQ datasets as exemplified in Appendix \ref{appendix:prompt_example}.

In the $k$-th sampling process for the $i$-th question $\boldsymbol{x}_i$, we employ each few-shot exemplar $\boldsymbol{p}_{k}\in \mathcal P$ with the question $\boldsymbol x_i$ to the LLM to generate the $k$-th response ${\boldsymbol{y}_i}^{(k)}$.
By taking $K$ times of the sampling process, we can obtain an answer set 
% $\boldsymbol Y_i=\left \{ \boldsymbol{y}_i^{(k)}|k \in \left \{1, \dots, K \right \} \right \}$ 
$\boldsymbol Y_i={\left \{ {\boldsymbol{y}_i}^{(k)} \right \}}_{k=1}^{K}$ 
to $\boldsymbol{x}_i$.
We set the labels 
% $\boldsymbol Z_i = \left \{ z^{(k)}_{i}|k \in \left \{1, \dots, K \right \} \right \}$ 
$\boldsymbol Z_i = {\left \{ {z_{i}}^{(k)} \right \}}_{k=1}^{K}$ 
by comparing each generated answer ${\boldsymbol{y}_{i}}^{(k)}$ with the ground-truth $\boldsymbol{\hat y}_i$ to indicate the correctness
(${z_{i}}^{(k)}\in\left \{ 0,1 \right \}$, 1 for \textit{True} and 0 for \textit{False}).
We collect and format the data in $(\boldsymbol x_i, \boldsymbol Y_i, \boldsymbol Z_i, \boldsymbol {\hat y}_i)$ in an extended dataset and calculate the uncertainty measures subsequently.
Note that since fine-tuning LLMs on unknown knowledge will encourage hallucinations \citep{zhang2024rtuning, gekhman-etal-2024-fine}, we revise the ground-truth answer to the question with ${z_{i}}^{(k)}=0, \forall {z_{i}}^{(k)}\in \boldsymbol Z_i$ to ``\textit{Sorry, I don't know.}'' to teach LLMs to refuse the questions beyond their knowledge \citep{zhang2024rtuning}.

\subsubsection{Uncertainty Measures}
\label{sssec:uncer}

In order to quantify the knowledge boundaries, we can leverage some uncertainty estimation methods.
The knowledge boundary of LLMs in this work is defined in two aspects.
The first involves the prior judgment to a question $\boldsymbol{x}_i$ regardless of the answers \citep{ren2023investigating} which indicates the certainty level of $\boldsymbol{x}_i$.
The second entails the dispersion measure to the distribution of the generated responses in $\boldsymbol Y_i$ to $\boldsymbol{x}_i$.
% We represent knowledge boundaries in two aspects, including the certainty level to the $i$-th question $\boldsymbol{x}_i$ and the distribution dispersion of the generated answers in $\boldsymbol Y_i$.
Accordingly, we adopt accuracy-based confidence \citep{xiong2024can} and semantic entropy \citep{kuhn2023semantic} to jointly determine and represent the actual knowledge boundary information.

\paragraph{Accuracy-based Confidence}
A natural idea of aggregating varied responses is to measure the accuracy among the candidate outputs to denote confidence scores \citep{manakul-etal-2023-selfcheckgpt,xiong2024can}.
Given a question $\boldsymbol{x}_i$, the accuracy of candidate responses in $\boldsymbol Y_i$ by comparing with the ground-truth answer $\boldsymbol{\hat y}_i$ serves as the confidence score $c_i$, computed as follows.

\vspace{-1em}
\begin{align}
    \label{eq:conf}
    c_i=\mathsf{Conf}(\boldsymbol{x}_i) = \frac{1}{K}\sum_{k=1}^{K}\mathds{1}\left (\boldsymbol{\tilde y}_i={\boldsymbol{y}_i}^{(k)} \right )
\end{align}

\paragraph{Semantic Entropy}
Due to the variable length and semantically equivalent generated sequences in sentence-level output spaces, \citet{kuhn2023semantic} proposes semantic entropy to capture uncertainty on the semantic level to quantify the degree of dispersion of sentence meanings.
The semantic entropy $e_i$ given $\boldsymbol x_i$ and $\boldsymbol Y_i$ is calculated as 

\vspace{-1em}
\begin{align}
    \label{eq:se}
    p(s|\boldsymbol{x}_i)&=\frac{1}{K}\sum_{k=1}^K \mathds{1}\left [{\boldsymbol{y}_i}^{(k)}\in s \right ] \\
    e_i=\mathsf{SE}(\boldsymbol{x}_i) &= -\sum_{s} p(s|\boldsymbol{x}_i)\log p(s|\boldsymbol{x}_i)
\end{align}
where $s$ denotes a set of sentences in semantic equivalent space.
As illustrated in Fig. \ref{fig:known}, semantic entropy is calculated by clustering semantically equivalent responses, as a measure to quantify the dispersion of generations to confirm the correct answer despite the low confidence, which will be further analyzed with the experimental results in Sec. \ref{ssec:measure}.
We calculate the confidence score and semantic entropy for both known and unknown questions.
Then we update a \textsc{UAlign} dataset $\mathcal D$ by formatting the $i$-th sample in $(\boldsymbol x_i, \boldsymbol Y_i, \boldsymbol Z_i, \boldsymbol {\hat y}_i, c_i, e_i)$.

\begin{figure}[t!]
  \centering
  \includegraphics[width=0.45\textwidth]{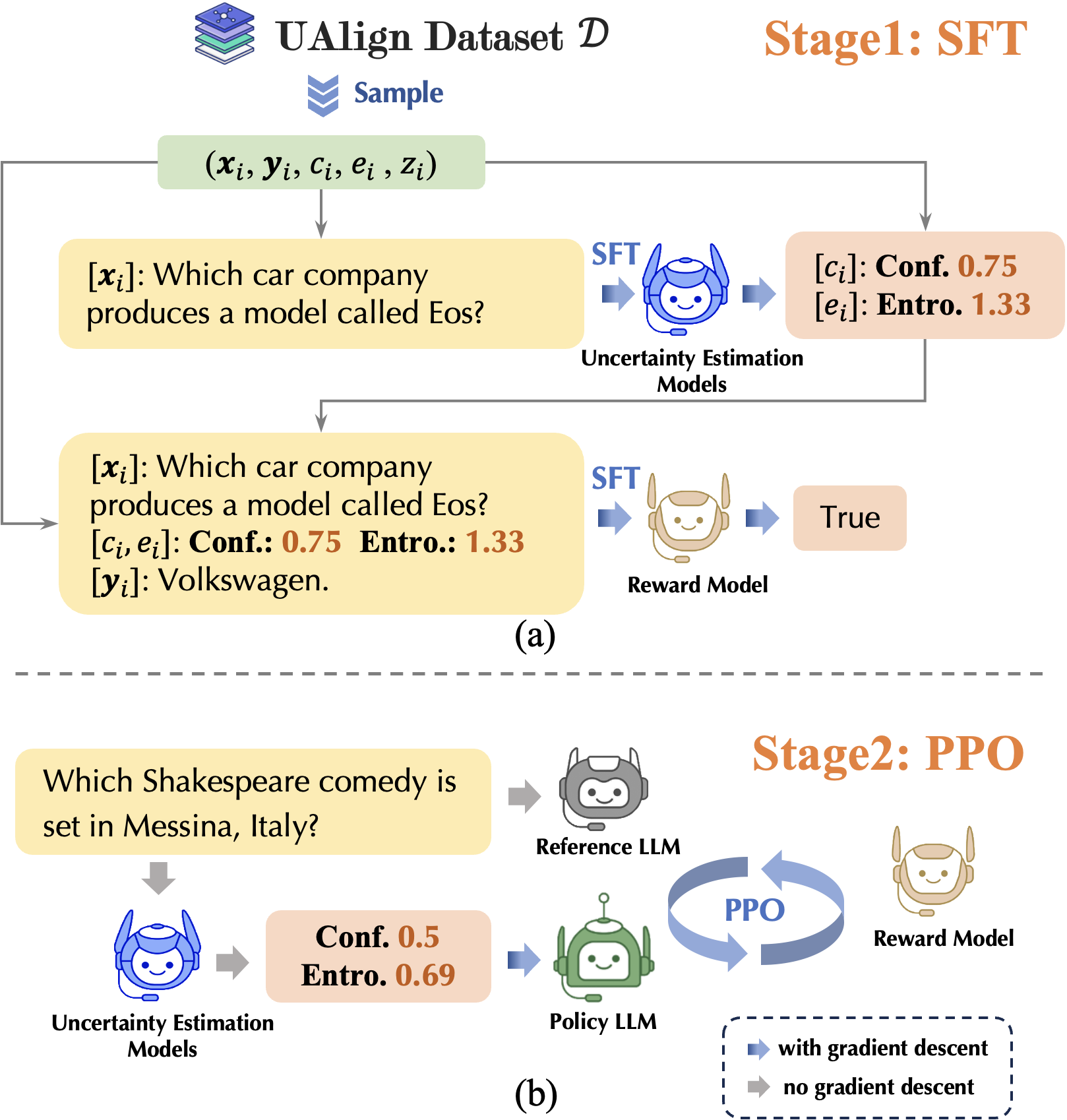}
    \caption{Illustration of (a) SFT and (b) PPO alignment processes of \textsc{UAlign} framework.
    Note that for simplicity, we only present one estimation model in the figure but there are actually two.}
  \label{fig:align}
\end{figure}

\subsection{\textsc{UAlign} Training Process}
\label{ssec:align}

% In this section, we demonstrate the training processes with the obtained {CUAlign} dataset $\mathcal D$, which is divided into a \textsc{CUAlign} SFT stage and a \textsc{CUAlign} PPO stage as shown in Fig. \ref{fig:align}.

\begin{algorithm}[t!]
    \caption{\textsc{UAlign} Training Algorithm}
    \label{algorithm}
    \normalsize
    \begin{algorithmic}[1]
         \State \textbf{Input:} \textsc{UAlign} dataset $\mathcal D$, uncertainty models $\tau~\mu$, reward model $\theta$, initial policy $\pi_o$.
        \State \textbf{Output:} Optimized policy $\pi_{\theta}$.
        \State \textbf{Stage 1: \textsc{UAlign} SFT}
        \State Train uncertainty models $\tau~\mu$ on $\mathcal{D}$ to predict $c_i, e_i$ by feeding $\boldsymbol x_i$ using Eq. \ref{eq:uncer_conf} and \ref{eq:uncer_entro}.
        \State Train reward model $\theta$ on $\mathcal D$ to predict $z_i$ by feeding $\boldsymbol x_i, c_i, e_i, {\boldsymbol y_i}^{(k)}$ using Eq. \ref{eq:reward}.
        \State \textbf{Stage 2: \textsc{UAlign} PPO}
        \State Collect reward $r$ including the reward signal $r_1$ by $\theta$ and \textrm{KL}-penalty $r_2$ between policy $\pi_{\theta}$ and initial policy $\pi_o$ as Eq. \ref{eq:ppo}.
        \State Update policy $\pi_\theta$ using the collected reward $r$.
    \end{algorithmic}
\end{algorithm}

% \begin{algorithm}
%     \caption{Reinforcement Learning from Human Feedback (RLHF)}
%     \begin{algorithmic}[1]
%         \State \textbf{Input:} Initial policy $\pi_\theta$, reward model $R_\phi$, 
%         number of feedback samples $N$, number of training iterations $T$
        
%         \State \textbf{Output:} Optimized policy $\pi_\theta^*$

%         \For{each iteration $t = 1, \ldots, T$}
%             \State \textbf{Step 1: Collect Human Feedback}
%             \For{each sample $i = 1, \ldots, N$}
%                 \State Generate a sample $(s_i, a_i)$ using policy $\pi_\theta$
%                 \State Get human feedback $f_i$ for the sample $(s_i, a_i)$
%                 \State Store the feedback in a dataset $\mathcal{D}$
%             \EndFor
            
%             \State \textbf{Step 2: Train Reward Model}
%             \State Train reward model $R_\phi$ using dataset $\mathcal{D}$ to predict human feedback:
%             \[
%             \phi^* = \arg\max_\phi \sum_{(s_i, a_i, f_i) \in \mathcal{D}} \log P(f_i | s_i, a_i; \phi)
%             \]

%             \State \textbf{Step 3: Optimize Policy}
%             \State Using the reward model $R_\phi$, optimize the policy $\pi_\theta$ with reinforcement learning (e.g., PPO):
%             \For{each epoch $e = 1, \ldots, E$}
%                 \State Collect experiences $(s_t, a_t, r_t)$ using policy $\pi_\theta$ where $r_t = R_\phi(s_t, a_t)$
%                 \State Update policy parameters $\theta$ using the collected experiences
%             \EndFor
%         \EndFor
%     \end{algorithmic}
% \end{algorithm}

\subsubsection{\textsc{UAlign SFT}: Uncertainty Estimation and Reward Models Training}
\label{sec:sft}

As presented in Fig. \ref{fig:align} (a) and Algorithm \ref{algorithm}, given dataset $\mathcal D$, \textsc{UAlign} SFT is to train uncertainty estimation models to explicitly learn the two estimations given specific questions.
Uncertainty estimation models of $\tau$ and $\mu$ are utilized to predict the confidence score and semantic entropy respectively, which are continuously used to train a reward model.
When training $\tau$ and $\mu$, we only feed a question $\boldsymbol x_i$ to the models to generate two uncertainty estimations.
The training objectives are to minimize the cross-entropy losses $\mathcal{L}_\tau$ and $\mathcal{L}_\mu$ as

% \vspace{-2em}
% \begin{align}
%     \arg&\min_{\tau}\mathcal{L}_\mathsf{Conf}, \arg\min_{\mu}\mathcal{L}_{\mathsf{SE}}\nonumber
% \end{align}

\vspace{-1em}
\begin{align}
\label{eq:uncer_conf}
    % &\arg\min_{\tau} \mathcal{L}_{\mathsf{Conf}, \mathcal{L}_{\mathsf{Conf}=
    \arg&\min_{\tau}\mathcal{L}_\tau, \arg\min_{\mu}\mathcal{L}_\mu,\nonumber\\
    % \vspace{-2em}
    \mathcal{L}_\tau&=-\mathbb{E}_{(\boldsymbol{x}_i,c_i)\sim\mathcal D}\left [ \log p_{\tau}(c_i|\boldsymbol{x}_i) \right ] \\
    \label{eq:uncer_entro}
    \mathcal{L}_\mu&=-\mathbb{E}_{(\boldsymbol{x}_i,e_i)\sim\mathcal D}\left [ \log p_{\mu}(e_i|\boldsymbol{x}_i) \right ]
    % \mathcal{L}_{\mathsf{Conf}} = -\log p_{\tau}(c_i|\boldsymbol{x}_i)\nonumber \\
    % &\arg\min_{\tau}\mathbb{E}_{(\boldsymbol{x}_i,c_i)\sim\mathcal D}\left [ \mathcal{L}_{\mathsf{Conf}} \right ],
    % \mathcal{L}_{\mathsf{Conf}} = -\log p_{\tau}(c_i|\boldsymbol{x}_i)\nonumber \\
    % \mathcal{L}_{\mathsf{SE}} = -\log p(e_i|\boldsymbol{x}_i)
    % \arg\max_{\mu}\mathbb{E}_{\boldsymbol{x}_i,c_i\in\mathcal D}\left [ \log p_{\mu}(c_i|\boldsymbol{x}_i) \right ]
    % &\arg\min_{\mu}\mathbb{E}_{(\boldsymbol{x}_i,e_i)\sim\mathcal D}\left [ \mathcal{L}_{\mathsf{SE}} \right ],
    % \mathcal{L}_{\mathsf{SE}} = -\log p_{\mu}(e_i|\boldsymbol{x}_i)
\end{align}
where the models can explicitly learn and express the uncertainty estimations which represent more accurate knowledge boundary information.

Subsequently, the reward model is introduced as a binary evaluator to determine if a generated answer ${\boldsymbol y_i}^{(k)}\in \boldsymbol Y_i$ is correctly conditioned on the question $\boldsymbol{x}_i$, confidence $c_i$, and entropy $e_i$.
Both $c_i$ and $e_i$ are explicitly used as additional auxiliary features to improve the accuracy of the reward model.
The binary cross-entropy loss $\mathcal{L}_{\theta}$ for the reward model $\theta$ is minimized as follows.

\vspace{-1em}
\begin{align}
\label{eq:reward}
    \arg&\min_{\theta}\mathcal{L}_{\theta},\mathcal{L}_{\theta} = -\mathbb{E}_{(\boldsymbol{x}_i,{\boldsymbol{y}_i}^{(k)},{z_i}^{(k)},c_i,e_i)\sim\mathcal D} [ {\mathcal{L}_{\theta}}^{(i)}  ]\nonumber\\
    % \arg&\min_{\theta}\mathbb{E}_{(\boldsymbol{x}_i,\boldsymbol{y}_i^{(k)},z_i^{(k)},c_i,e_i)\sim\mathcal D}\left [ \mathcal{L}_{\mathsf{Conf}} \right ], \nonumber\\
    {\mathcal{L}_{\theta}}^{(i)} &=  -{{z}_i}^{(k)}\log p_\theta({{z}_i}^{(k)}|\boldsymbol{x}_i, c_i, e_i, {\boldsymbol y_i}^{(k)}) \nonumber\\
    -&(1-{{z}_i}^{(k)})\log (1-p_\theta({{z}_i}^{(k)}|\boldsymbol{x}_i, c_i, e_i, {\boldsymbol y_i}^{(k)}))
\end{align}

\subsubsection{\textsc{UAlign} PPO: Policy Model Training}
\label{sec:ppo}

The \textsc{UAlign} PPO is to elicit the LLM's factual expressions to a question with the uncertainty measures using obtained models.
Inspired by the progress of reinforcement learning from human feedback (RLHF) technique \cite{ouyang2022training,ziegler2019fine}, we employ proximal policy optimization (PPO) \citep{schulman2017proximal} for LLM optimization with the reward model $\theta$.
As illustrated in Fig. \ref{fig:align} (b), the LLM to be optimized is used as the policy $\pi_{\theta}$.
During this phase, we iteratively feed the question $\boldsymbol x$, and the predicted confidence $c$ and entropy $e$ to both the policy $\pi_{\theta}$ and the reference $\pi_o$, and the reward function $r$ will facilitate reliable expressions of $\boldsymbol y$ of the policy model $\pi_{\theta}$.
Model update details are further specified in Appendix \ref{appendix:ppo}.
% Only the policy model is optimized during training while the uncertainty and reward models are frozen as specified in Appendix \ref{}.
The training objective is to maximize the following reward function $r$ as

\vspace{-1em}
\begin{align}
\label{eq:ppo}
    \arg\max_{\pi_{\theta}} \mathbb{E}_{\boldsymbol{x}\sim\mathcal D,c\sim {\tau}(\boldsymbol{x}),e\sim {\mu}(\boldsymbol{x}),\boldsymbol{y}\sim \pi_{\theta}(\boldsymbol{x},c,e)}\left [ r \right ]\nonumber\\
    r=\underbrace{\theta(\boldsymbol{x},\boldsymbol{y},c,e)}_{r_1}-\beta\underbrace{ \mathrm{KL}[\pi_{\theta}(\boldsymbol{x},c,e)||\pi_o(\boldsymbol{x})]}_{r_2}
\end{align}
where the reward function $r$ contains a reward signal $r_1$ from $\theta$ and a $\mathrm{KL}$-penalty $r_2$ to make sure the generated answers $\boldsymbol y$ by policy $\pi_{\theta}$ don't diverge too much from the original policy $\pi_o$.
The hyper-parameter $\beta$ is the coefficient of $\mathrm{KL}$-penalty.

% \clearpage

\section{Experimental Setting}
\label{sec:exp}

% In this section, we first provide an overview of the
% benchmark datasets and the corresponding evaluation settings.
% Then the baseline models and the implementation details are presented in the following subsections, respectively.

\subsection{Datasets}
\label{sec:dataset}

The \textsc{UAlign} training set is comprised of three widely used knowledge-intensive QA datasets: \textbf{TriviaQA} (\textbf{TVQA}) \citep{joshi-etal-2017-triviaqa} which contains closed-book trivia QA pairs to gauge models’ factual knowledge, \textbf{SciQ} \citep{welbl2017sciq} requiring scientific professional knowledge, and \textbf{NQ-Open} \citep{kwiatkowski-etal-2019-natural} which is constructed by Google Search queries along with annotated short answers or documents.

For testing, we evaluate the in-domain (ID) performance on the corresponding validation/test sets and generalization on an out-of-domain (OOD) test set \textbf{LSQA} \citep{xue2024comprehensive} which contains multilingual language-specific QA pairs.
More dataset details and statistics are presented in Appendix \ref{appendix:dataset}.

\subsection{Evaluation Metrics}
\label{sec:eval}

To evaluate LLMs' reliability, we employ two metrics: \textit{Precision} (\textit{Prec.}) and \textit{Truthfulness}  (\textit{Truth.}). 
\textit{Precision} is defined as the proportion of correctly answered questions among all the known questions, representing LLMs' ability to accurately express their known factual knowledge.
\textit{Truthfulness} represents the proportion of the sum of correctly answered known and refused unknown questions among all questions, indicating LLMs' honesty level.
Details can be referred to Appendix \ref{appendix:prec_truth}.

To ascertain the correctness of the LLM-generated answer $\boldsymbol y$ with the ground truth $\boldsymbol {\hat y}$, we employ a string-matching approach.
Exact matching (EM) of $\boldsymbol y\equiv \boldsymbol {\hat y}$ always misjudges some correct answers with slight distinctions on such closed-book QA tasks.
Therefore, we replace EM with a variant of $\boldsymbol y\in \boldsymbol {\hat y} \vee \boldsymbol {\hat y} \in \boldsymbol y$ to evaluate the accuracy.
The specific illustrations of evaluation formulas and comparisons of several EM variants we tested with human evaluations are in Appendix \ref{appendix:accu}.

\subsection{Baselines}
\label{ssec:baseline}

We present several baselines in four categories below.
To clearly delineate the differences between our proposed method and other baselines, we have illustrated all methods in Fig. \ref{fig:baseline} in Appendix \ref{appendix:baseline}.

\paragraph{Prompt-based} We present two prompt-based baselines namely In-Context Learning (\textbf{ICL}), In-Context Learning with Refusal Examples (\textbf{ICL-IDK}), and In-Context Learning Chain-of-Thought (\textbf{ICL-CoT}) \citep{wei2022chain}.
The few-shot prompt templates are presented in Appendix \ref{appendix:prompt}.

\paragraph{SFT-based} We employ standard Supervised Fine-Tuning (\textbf{SFT}) by training an LLM to generate answers for all questions.
We also introduce \textbf{R-Tuning} \citep{zhang2024rtuning} which teaches LLM to refuse their unknown questions.

\paragraph{RL-based} Following {RLHF} technique \citep{ouyang2022training}, we first train a reward model to determine correctness by SFT.
Then we employ PPO to optimize the policy model with the reward model (\textbf{RL-PPO}).
We also introduce an advanced variant called reinforcement learning from knowledge feedback (\textbf{RLKF}) \citep{liang-etal-2024-learning} which leverages knowledge probing and consistency checking to train the reward model. 
Following \citet{zhang2024selfalignment,tian2024finetuning,lin2024flame}, we also construct the factuality preference dataset to conduct direct preference optimization (\textbf{RL-DPO}) to enhance the factuality of LLMs.

\paragraph{Inference-based} Another branch of work focuses on shifting the output distribution to improve factuality during inference. 
\citet{li2023inferencetime} (\textbf{ITI}) intervenes in the activations in attention heads to the ``truthfulness'' direction.
% \citet{chuang2024dola} (\textbf{DoLa}) surfaces factual knowledge by contrasting logits in different layers when decoding.

\subsection{Implementation Details}
\label{sec:implement}

Experiments are conducted on two LLMs: {\textbf{{Llama-3-8B}}} ({Llama-3}) \footnote{\href{https://huggingface.co/meta-llama/Meta-Llama-3-8B}{https://huggingface.co/meta-llama/Meta-Llama-3-8B}} \citep{llama3modelcard} and {\textbf{{Mistral-7B}}} ({Mistral}) \footnote{\href{https://huggingface.co/mistralai/Mistral-7B-v0.1}{https://huggingface.co/mistralai/Mistral-7B-v0.1}} \citep{jiang2023mistral7b}.
When preparing the \textsc{UAlign} dataset, we sample 10 responses for each question on $K=10$ different 1-shot prompts.
The sampling temperature $T$ is set to 0.2 to achieve a trade-off between the diversity and factuality of the answer set.
During training, all the LLMs are trained using LoRA \citep{hu2022lora} with rank $r=16$.
Both the uncertainty estimation models and the reward model utilize the vanilla LLM as their bases and are trained using LoRA with rank $r=4$.
ADAM parameter update is used in a mini-batch mode.
Uncertainty estimation models and the reward model are trained using SFT on the \textsc{UAlign} dataset.
The \textsc{UAlign} PPO algorithm and all the RL-based baselines are implemented by \texttt{trl}~\footnote{\href{https://github.com/huggingface/trl}{https://github.com/huggingface/trl}}.
All training hyper-parameters are presented in Appendix \ref{appendix:training}.
When decoding, the temperature is also set to 0.2 to be consistent with the sampling setting.
All the experiments are conducted on 4 $\times$ NVIDIA A100-40GB GPUs.

\section{Results and Analysis}
\label{sec:result}

\begin{table*}[t]
    \centering
    \footnotesize
    \resizebox{.96\textwidth}{!}
    {\begin{tabular}{lcccccccccc}
    \toprule
        \multirow{2}{*}{\bf Method} & \multicolumn{2}{c}{\bf TVQA (ID)} & \multicolumn{2}{c}{\bf SciQ (ID)} & \multicolumn{2}{c}{\bf NQ-Open (ID)} & \multicolumn{2}{c}{\bf Avg. (ID)} & \multicolumn{2}{c}{\bf LSQA (OOD)} \\
         & \it Prec. $\uparrow$ & \it Truth. $\uparrow$ & \it Prec. $\uparrow$ & \it Truth. $\uparrow$ & \it Prec. $\uparrow$ & \it Truth. $\uparrow$ & \it Prec. $\uparrow$ & \it Truth. $\uparrow$ & \it Prec. $\uparrow$ & \it Truth. $\uparrow$ \\
        \hline
        \hline
        \rowcolor{pearl}
        \multicolumn{11}{c}{\textbf{\textsl{Llama-3-8B}}} \\
        \hline
        \bf ICL & 76.15 & 56.55 & 70.43 & 44.30 & 50.28 & 20.11 & 65.62 & 40.32 & 77.35 & 52.98 \\
        \bf ICL-IDK & 69.17 & 54.10 & 68.36 & 43.00 & 45.43 & 20.72 & 60.98 & 39.27 & 66.67 & 50.24 \\
        \bf ICL-CoT & 66.68 & 53.37 & 72.34 & 45.90 & \cellcolor{platinum} \bf 57.34 & 23.60 & 65.45 & 40.95 & 73.96 & 49.37 \\
        \hdashline
        \bf SFT & 70.80 & 52.57 & 72.18 & 45.40 & 41.41 & 16.57 & 61.46 & 38.18 & 68.09 & 46.63 \\
        {\bf R-Tuning} & 72.93 & 55.44 & 71.38 & 44.90 & 47.81 & 18.12 & 64.04 & 39.48 & 71.54 & 52.15 \\
        \hdashline
        \bf RL-PPO & 76.32 & 55.19 & 75.70 & 45.80 & 54.07 & 24.19 & 68.03 & 41.72 & 72.18 & 48.43 \\
        \bf RL-DPO & 72.08 & 53.96 & 71.23 & 44.20 & 49.65 & 19.18 & 64.32 & 39.11 & 71.09 & 48.88 \\
        {\bf RLKF} & 77.12 & 56.07 & 72.36 & 44.90 & 54.86 & 22.15 & 68.11 & 41.04 & 74.95 & 52.46 \\
        \hdashline
        \bf ITI & 71.09 & 53.97 & 72.35 & 43.80 & 43.20 & 17.13 & 62.21 & 38.30 & 68.52 & 46.99 \\
        \hline
        {\bf \textsc{UAlign}} & \cellcolor{platinum} \bf 79.14 & 57.04 & \cellcolor{platinum} \bf 76.44 & \cellcolor{platinum} \bf 48.00 & 56.60 & 26.09 & \cellcolor{platinum} \bf 70.72 & 43.71 & \cellcolor{platinum} \bf 79.56 & \cellcolor{platinum} \bf 55.88 \\
        \ \ \ (w/o {\bf Conf.})
        & 74.13 & 54.45 & 74.05 & 45.00 & 54.19 & 23.60 & 67.45 & 41.01 & 74.25 & 52.06 \\
        \ \ \ (w/o {\bf Entro.})& 78.43 & \cellcolor{platinum} \bf 57.69 & 75.39 & 47.50 & 56.68 & \cellcolor{platinum} \bf 27.56 & 70.16 & \cellcolor{platinum} \bf 44.25 & 76.14 & 54.43 \\
        \hline
        \rowcolor{pearl}
        \multicolumn{11}{c}{\textbf{\textsl{Mistral-7B}}} \\
        \hline
        \bf ICL & 77.92 & 55.14 & 68.62 & 42.20 & 52.09 & 17.95 & 66.21 & 38.43 & 74.09 & 47.71 \\
        \bf ICL-IDK & 72.59 & 51.37 & 63.74 & 39.20 & 51.13 & 17.67 & 62.48 & 36.20 & 72.27 & 47.32 \\
        \bf ICL-CoT & 76.73 & 54.78 & 71.87 & 44.20 & \cellcolor{platinum} \bf 54.47 & 18.22 & 67.69 & 39.06 & \cellcolor{platinum} \bf 79.24 & 52.59 \\
        \hdashline
        \bf SFT & 74.57 & 54.77 & 65.85 & 42.50 & 50.82 & 14.42 & 63.74 & 37.08 & 68.33 & 44.00 \\
        {\bf R-Tuning} & 67.70 & 52.25 & 64.44 & 40.10 & 46.33 & 15.52 & 59.49 & 36.29 & 64.67 & 44.05 \\
        \hdashline
        \bf RL-PPO & 79.23 & 55.08 & 71.35 & 44.10 & 53.76 & 19.19 & 68.11 & 39.45 & 74.49 & 49.67 \\
        \bf RL-DPO & 72.20 & 52.98 & 66.44 & 41.80 & 50.95 & 16.42 & 63.19 & 37.06 & 67.82 & 43.77 \\
        {\bf RLKF} & 80.43 & 56.92 & 70.66 & 43.90 & 52.09 & 18.24 & 67.72 & 39.68 & 74.19 & 49.23 \\
        \hdashline
        \bf ITI & 74.65 & 55.16 & 66.90 & 44.90 & 51.12 & 16.68 & 64.22 & 38.91 & 67.73 & 46.20 \\
        \hline
        {\bf \textsc{UAlign}} & \cellcolor{platinum} \bf 82.10 & \cellcolor{platinum} \bf 59.05 & \cellcolor{platinum} \bf 73.21 & \cellcolor{platinum} \bf 46.70 & 54.17 & \cellcolor{platinum} \bf 19.64 & \cellcolor{platinum} \bf 70.82 & \cellcolor{platinum} \bf 41.79 & 76.29 & \cellcolor{platinum} \bf 52.89 \\
        \ \ \ (w/o {\bf Conf.}) & 76.44 & 55.13 & 69.84 & 43.50 & 50.30 & 17.88 & 65.52 & 38.83 & 73.15 & 47.06 \\
        \ \ \ (w/o {\bf Entro.}) & 80.18 & 57.64 & 72.90 & 45.60 & 52.21 & 18.44 & 68.43 & 40.56 & 75.34 & 50.15 \\
        \hline
    \bottomrule
    \end{tabular}}
    \caption{Experiments of {Precision} (\textit{Prec.}) and {Truthfulness} (\textit{Truth.}) on four datasets on {{Llama-3}} and {{Mistral}}.}
\label{table:main_exp}
\end{table*}

\subsection{Main Experimental Results}

We present the results of \textsc{UAlign} and several baselines on three ID and one OOD test sets as shown in Table \ref{table:main_exp}.
Several findings are listed below.

\paragraph{Reliability}
% As shown in Table \ref{table:main}, several trends can be found:
Significant improvements are consistently achieved on diverse datasets using the proposed \textsc{UAlign} framework over other baseline methods on both Llama-3 and Mistral.
We highlight the supreme Precision and Truthfulness performance using grey highlights among the all baselines of each column in Table \ref{table:main_exp}.
The core idea of our \textsc{UAlign} framework is the utilization of uncertainty estimation models.
Compared with the most relevant baselines of RL-PPO and RLKF, both the reward model and policy model in \textsc{UAlign} generate predictions and responses conditioned on uncertainty estimations regarding the knowledge boundaries to questions, thereby yielding better reliability performance.
It can be attributed that by explicitly appending uncertainty measures following the question, LLMs can be assisted to elicit more accurate responses based on intrinsic knowledge boundary representations.

\paragraph{Generalization}
We also introduced an OOD test set to assess the generalization capability of the \textsc{UAlign} method.
The results in Table \ref{table:main_exp} indicate that most training-based baselines (SFT, RL, Inference) are unstable and result in performance decreasing compared with prompt-based baselines when generalizing on the OOD test set.
However, comparable reliability performances are obtained on two LLMs using the proposed \textsc{UAlign} in comparison with prompt-based methods, demonstrating strong generalization capability.

\begin{table}[t!]
    \centering
    \footnotesize
    \resizebox{.45\textwidth}{!}
    {\begin{tabular}{ccccccccccc}
    \toprule
        \multirow{2}{*}{\textbf{Conf.}} & \multirow{2}{*}{\textbf{Entro.}} & \multicolumn{3}{c}{\bf ID} & \bf OOD \\
         & & \bf TVQA & \bf SciQ & \bf NQ-Open & \bf LSQA \\
        \hline
        \hline
        \rowcolor{pearl}
        \multicolumn{6}{c}{\textbf{\textsl{Llama-3-8B}}} \\
        \hline
        \xmark & \xmark & 82.31 & 79.00 & 67.45 & 70.12 \\
        \cmark & \xmark & 85.41 & 84.30 & 70.37 & \cellcolor{platinum} \bf 75.09 \\
        \xmark & \cmark & 82.05 & 77.90 & 67.85 & 70.40 \\
        \cmark & \cmark & \cellcolor{platinum} \bf 86.73 & \cellcolor{platinum} \bf 86.40 & \cellcolor{platinum} \bf 72.00 & 74.59 \\
        \hline
        \rowcolor{pearl}
        \multicolumn{6}{c}{\textbf{\textsl{Mistral-7B}}} \\
        \hline
        \xmark & \xmark & 84.53 & 77.30 & 65.24 & 68.31 \\
        \cmark & \xmark & 86.80 & 79.50 & 72.10 & 72.95 \\
        \xmark & \cmark & 85.24 & 74.60 & 66.64 & 71.22 \\
        \cmark & \cmark & \cellcolor{platinum} \bf 88.06 & \cellcolor{platinum} \bf 79.80 & \cellcolor{platinum} \bf 75.14 & \cellcolor{platinum} \bf 73.61 \\
    \bottomrule
    \end{tabular}}
    \caption{Accuracy of reward model varying different uses of uncertainty measures {Conf.} and {Entro.} in \textsc{UAlign} dataset on {{Llama-3}} and {{Mistral}}.}
\label{table:reward}
\end{table}

\begin{figure}[!ht]
  \centering
  \includegraphics[width=0.46\textwidth]{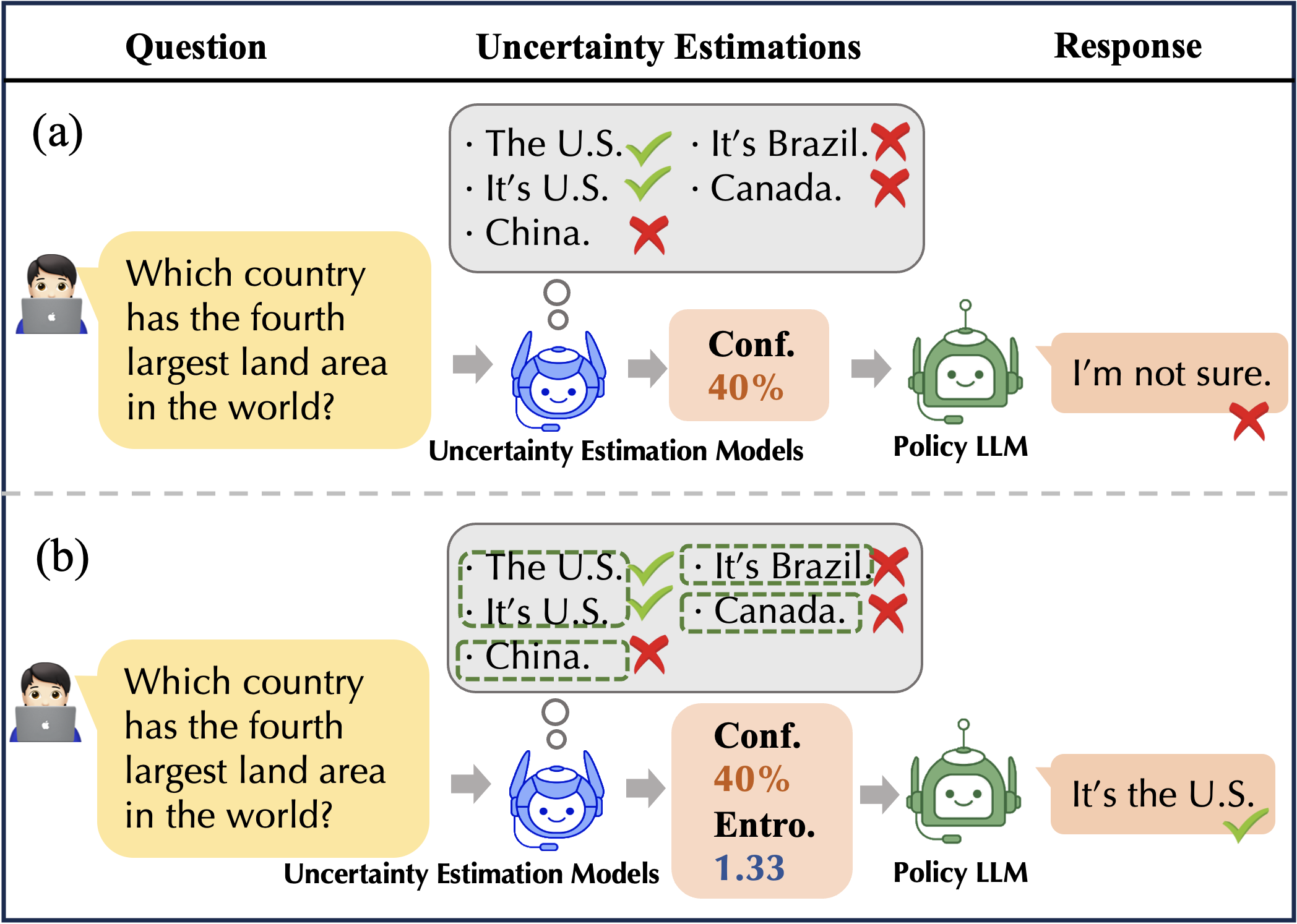}
    \caption{Illustration of the effects of different uses of uncertainty estimations under varying knowledge boundaries perceived by LLMs.}
  \label{fig:known}
\end{figure}

\subsection{Effects of Uncertainty Estimation Models}
\label{ssec:measure}

\paragraph{Setting}
To investigate the effects of introducing uncertainty estimations as input features to reward models, we report the accuracy of reward models that vary in different uses of two measures on ID and OOD tasks.
The reward models are trained on the \textsc{UAlign} dataset on both Llama-3 and Mistral.

\paragraph{Results} 
As in Table \ref{table:reward}, we present the results of the accuracy of reward models.
Significant accuracy improvements of reward models are obtained that predominantly benefit from the use of confidence scores across both ID and OOD test sets on two LLMs, validating the effectiveness of our proposed \textsc{UAlign} framework.
The isolated use of semantic entropy does not guarantee a stable improvement but may even lead to a performance decrease on some test sets.
However, when semantic entropy is employed in combination with confidence measures, it can facilitate further enhancements, achieving optimal results across most test sets as highlighted grey cells for two LLMs.

\paragraph{Analysis}
In the \textsc{UAlign} framework, both confidence score and semantic entropy are introduced to quantify the intrinsic knowledge boundary of LLMs to questions.
The explicit introduction of the knowledge boundary representations in prompts can be regarded as the added thinking step like CoT.
The combined use of confidence and semantic entropy can achieve supreme prediction performance in Table \ref{table:reward}.
% It can be assumably attributed that when leveraging both certainty and uncertainty estimations, they can complement each other for more accurate knowledge boundary representations to improve the accuracy of the reward models.
We illustrate the mechanism as follows.

As demonstrated in Fig. \ref{fig:known} (a), by sampling multiple responses to a question, we can approximate LLM's intrinsic knowledge boundary, where the certainty level of the answer ``\textit{The U.S.}'' is 40\%.
In previous work \citep{zhang2024rtuning} which only considers the confidence level, the correct answer that the LLM knows but is not sure will be discarded and the LLM will refuse to answer.
However, as in Fig. \ref{fig:known} (b), the LLM can perceive that even though its certainty level to the correct answer is low, other answers are more uncertain and the dispersion level of answers is relatively high which is quantified by semantic entropy.
After \textsc{UAlign} PPO training, the ability to generate correct answers conditioned on questions and estimations is well enhanced.
As a result, the correct but unsure knowledge will be elicited in the responses.

% \subsection{Ablation Study}

\begin{figure}[!ht]
  \centering
  \includegraphics[width=0.34\textwidth]{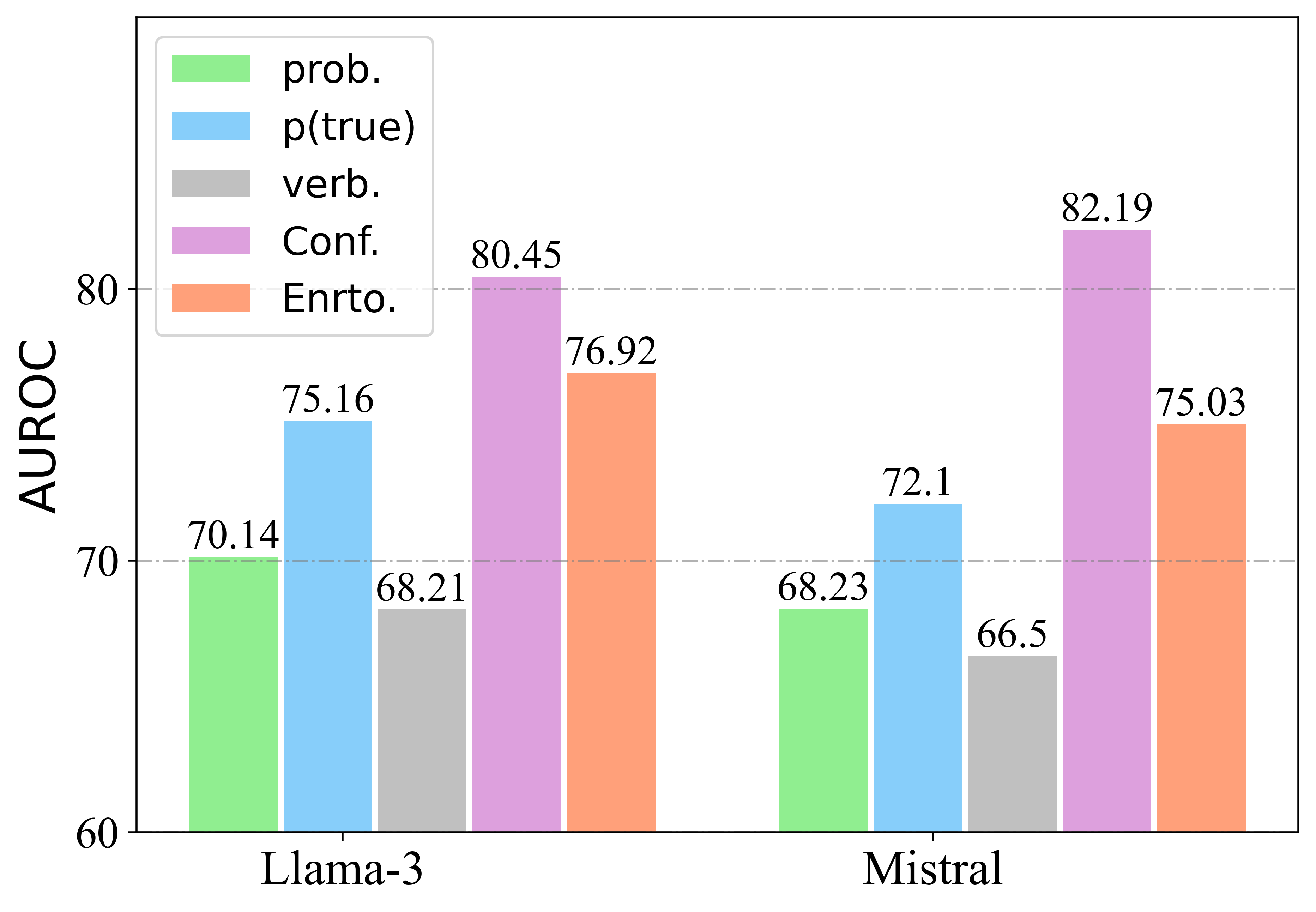}
    \caption{Results of AUORC$\uparrow$ of several uncertainty estimation methods on TVQA using Llama-3 and Mistral.}
  \label{fig:auroc_tvqa}
\end{figure}

\subsection{Reliability of Uncertainty Estimations}

\paragraph{Setting}
Evaluating the performance of confidence score and semantic entropy is essential to the \textsc{UAlign} method.
We present the AUROC (Detailed in Appendix \ref{appendix:evaluate}) results of two estimations in comparison with three confidence/uncertainty estimation methods (one probability-based method (Prob.), two prompt-based methods including p(True) and verbalized (Verb.) as illustrated in Fig. \ref{fig:uncer}) on TriviaQA on two LLMs.
Results on other datasets are remained in Appendix \ref{appendix:exp}.
Details of baseline estimation baselines are presented in Sec. \ref{sec:related}, Appendix \ref{appendix:relate}, and Fig. \ref{fig:uncer}.

\paragraph{Results}
In Fig. \ref{fig:auroc_tvqa}, both the confidence and entropy prediction consistently outperform other baseline uncertainty estimation methods.
Optimal AUROC performances are obtained using confidence on both Llama-3 (80.45) and Mistral (82.19).

\paragraph{Analysis}
After \textsc{UAlign} SFT stage, the uncertainty estimation models are converged on the \textsc{UAlign} dataset to predict both confidence and entropy, indicating the models possess the ability to predict the two measures.
Practically, our utilized confidence and semantic entropy incorporate the advantages of both sampling- and training-based uncertainty estimations.
Multiple sampling can better approximate the actual knowledge boundaries of LLMs, while the training-based approach enables the LLMs to learn to perceive their intrinsic knowledge boundaries.
Compared to other baselines that suffer from overconfidence issues with low AUROC scores, our utilized methods yield more reliable estimates, thereby ensuring improved performance for both the reward model and the policy model in the following stages.

\begin{figure}[!ht]
  \centering
  \includegraphics[width=0.46\textwidth]{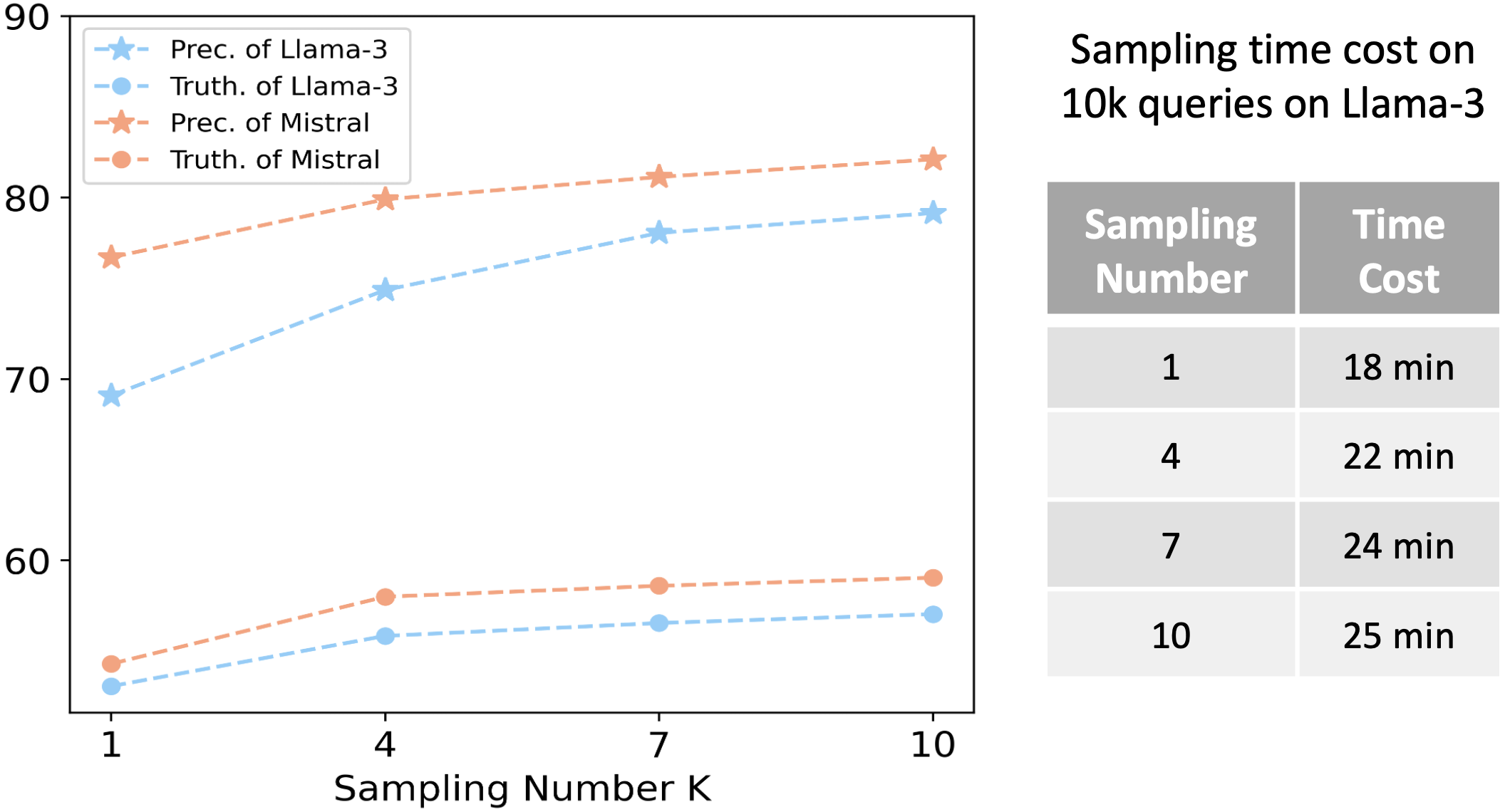}
    \caption{Experiments of \textit{Prec.}, \textit{Truth.} (left), and time costs (right) of various sampling number $K$ of 1, 4, 7, and 10 on TVQA on both Llama-3 and Mistral.}
  \label{fig:sample}
\end{figure}

\subsection{Effects of Sampling Number}
\label{ssec:sample_num}

\paragraph{Setting}
The sampling number $K$ is a crucial hyper-parameter in the \textsc{UAlign} method. 
Different values of $K$ can significantly affect the precision of the knowledge boundary measurements.
To evaluate the effects, we compare performances using various $K$ of 1, 4, 7, and 10.
Experimental results on TVQA are presented in Fig. \ref{fig:sample}.

\paragraph{Findings}
Results in Fig. \ref{fig:sample} indicate that when using small sampling numbers, increasing $K$ leads to significant improvements in both \textit{Prec.} and \textit{Truth.}.
However, as $K$ increases, the reliability improvement tends to plateau, exhibiting convergence.
Therefore, we opt $K=10$ as the optimal setting and don't experiment using larger $K$.
% Since further increasing $K$ requires extra costs, we discard conducting experiments with larger $K$.

We also report the sampling time costs to construct the training set in Fig. \ref{fig:sample}, and further specify cost analysis of \textsc{UAlign} dataset construction and inference time in Sec.~\ref{ssec:cost}, Appendix \ref{appendix:train_cost} and \ref{appendix:infer_cost} respectively.
We showcase that with various acceleration and quantization methods, time costs of \textsc{UAlign} can be significantly reduced when scaling to larger models or datasets, exhibiting both efficiency and efficacy.

\paragraph{Analysis}
The results in Fig. \ref{fig:sample} demonstrate that while the sampling number $K$ increases linearly, the performance improvements are non-linear. 
This may be attributed to utilizing non-linear metrics, or it could suggest that $K=10$ can approximate the actual knowledge boundaries, resulting in a gradual slowdown in performance gains.
Consequently, setting $K$ to 10 in this work makes a trade-off between performance gains and computation expense.

\subsection{Computation Cost Analysis}
\label{ssec:cost}

We test the time costs to construct the UAlign dataset $\mathcal D$ in different sampling numbers.
We present different sampling time costs on 10000 QA samples on Llama-3-8B on 4×40G A100 GPUs as presented in Fig. \ref{fig:sample}.
Results demonstrate the efficiency of our proposed \textsc{UAlign} method even though scaling on larger models.
The relatively low computation costs when sampling can be attributed that experiments are conducted on knowledge QA datasets. 
The answer spans are entity-level, and each answer only needs to generate a few tokens.
Since the output form is relatively simple, sampling ten times is sufficient and cost-saving to accurately fit the knowledge boundaries as in Sec. \ref{ssec:sample_num}.

Furthermore, Test Time Scaling Law \citep{snell2024scalingllmtesttimecompute} has attracted much attention recently, which proposed to consider allocating more computation resources in inference to generate high-quality responses.
These LLMs' self-generated data can be further used for LLM training to self-improve LLMs \citep{gulcehre2023reinforcedselftrainingrestlanguage}.
Many works validate that incorporating multiply-sampled data on LLMs in inference can benefit LLMs for further improvements, which is a new trend for LLM training. 
In this way, our proposed \textsc{UAlign} provides a novel insight into the test time scaling law to represent knowledge boundaries by calculating uncertainty estimations on the sampled responses, and further explicitly leverages the uncertainty estimations for factuality alignment, heralding a promising view of the test time scaling law.
Although a few additional computation costs are required, our proposed \textsc{UAlign} is still efficient to be utilized practically with significant reliability improvements.

% \clearpage

\section{Related Works}
\label{sec:related}

\paragraph{Knowledge Boundary}

Previous works investigate the knowledge boundary \citep{yin2024benchmarking,wang2024selfdc} to identify the known level of a knowledge piece of LLMs by quantifying uncertainty estimations like output consistency \citep{cheng2024aiassistants}, prompting methods \citep{ren2023investigating} or knowledge probing \citep{ji-etal-2024-llm}.
Generally, knowledge boundary measures derive from uncertainty estimations.

\paragraph{Uncertainty Estimation for LLMs}
We categorize uncertainty estimation methods on LLMs into four classes as illustrated in Figure \ref{fig:uncer}.
\textit{\circledone~Likelihood-based methods} \citet{vazhentsev-etal-2023-efficient} directly quantify sentence uncertainty over token probabilities;
\textit{\circledtwo~Prompting-based methods} instruct LLMs to express uncertainty in words \citep{lin2022teaching,xiong2024can} or to self-evaluate its correctness on $p(\mathrm{True})$ \citep{kadavath2022language};
\textit{\circledthree~Sampling-based methods} aggregate sampled responses to calculate consistency \citep{xiong2024can} or semantic entropy 
 \citep{kuhn2023semantic};
\textit{\circledfour~Training-based methods} \citep{lin2022teaching} propose to train LLMs to improve linguistic uncertainty expressions.

\paragraph{Factuality Alignment}

LLM alignment is to guide human preference through Reinforcement Learning from Human Feedback (RLHF) \citep{ouyang2022training,bai2022training}.
Distinct from recent studies that apply RL to improve LLMs’ factuality \citep{zhang2024selfalignment,lin2024flame,liang-etal-2024-learning,xu2024rejection},
this work improves LLMs' reliability by explicitly leveraging the uncertainty estimations for LLM alignment.

\paragraph{}
Due to the space limitation, detailed investigations of related works are shown in Appendix \ref{appendix:relate}.

\section{Conclusion}
\label{sec:conclu}

In this paper, we present a \textsc{UAlign} framework to explicitly leverage uncertainty estimations to elicit LLMs to accurately express factual knowledge that LLMs cannot constantly answer correctly due to ambiguous knowledge boundaries.
We introduce the dataset preparation process and \textsc{UAlign} training strategies of factuality alignment by incorporating uncertainty estimations of the confidence score and semantic entropy as input features into prompts.
Experiments on several knowledge QA tasks affirm the efficacy of \textsc{UAlign} to enhance the LLMs' reliability and generalizability, demonstrating significant improvements over various baselines.

% \clearpage

\section*{Limitations}
\label{sec:limit}

The limitations and future work of this study are listed as follows:

\paragraph{Task Expansion}
The dataset used in this paper is solely based on factual knowledge QA tasks, with a simple and fixed template and response format.
However, the \textsc{UAlign} methodology has not been further validated on other factual knowledge-based tasks such as open-form instruction-following tasks, long-form generation like biography, or even knowledge reasoning tasks, where the uncertainty estimations remain challenging. In future works, we plan to extend the \textsc{UAlign} framework to open-ended generation tasks to enhance the LLMs' factual expressions.

\paragraph{Computational Cost}
The current method for constructing the {\textsc{UAlign}} dataset relies on multiple samplings, requiring additional computational cost that linearly increases with the number of sampling instances $K$ and a higher number of samplings is preferable to accurately approximate the knowledge boundaries.
As we have adopted a range of acceleration and quantization methods to reduce the time cost during both constructing the dataset and inference as presented in Appendix \ref{appendix:train_cost} and \ref{appendix:infer_cost}, there remains potential for exploration to further alleviate computational resources requirements.

\section*{Ethical Statement}
\label{sec:ethical}

In this paper, three evaluators are employed to annotate the correctness of four EM variants on selected samples, which aims to select the optimal EM variant to evaluate the correctness of the generated answer and the ground-truth label as presented in Appendix \ref{appendix:accu}.
All the evaluators are M.Phil. or Ph.D. students possessing sufficient expertise to carry out the evaluation. 
We meticulously adhered to legal and ethical standards throughout the human evaluation process, prioritizing privacy and obtaining informed consent.
Evaluators were furnished with comprehensive details regarding the study’s objectives, data collection methodologies, and associated risks or benefits. 
They were afforded the opportunity to seek clarifications and voluntarily provide consent before their involvement.
All the human evaluation results were exclusively utilized for research purposes.

\section*{Acknowledgments}

This work was partially supported by Hong Kong RGC GRF No. 14206324, CUHK direct grant No. 4055209, and CUHK Knowledge Transfer Project Fund No. KPF23GWP20.

\bibliography{custom}

% \clearpage

\appendix

% \begin{figure*}[htbp]
% \centering
% \begin{spacing}{0.15} % 设置行间距
% \setlength\tabcolsep{1pt} % 设置列间距
% \begin{tabular}{|c|c|}
%     \toprule \\
%     \bf Confidence \& Uncertainty Estimation Methods on LLMs & \bf Disadvantages \\
%     \hline \\
%     \begin{minipage}[b]{1.4\columnwidth}
%     		\centering
%     		\raisebox{-.5\height}{\includegraphics[width=0.8\linewidth]{figs/uncer-likelihood.png}} 
%     \end{minipage} & 1 \\
%     \hline \\
%         \begin{minipage}[b]{1.2\columnwidth}
%     		\centering
%     		\raisebox{-.5\height}{\includegraphics[width=\linewidth]{figs/uncer-likelihood.png}} 
%     \end{minipage} & 1 \\
%     \hline \\
%         \begin{minipage}[b]{1.2\columnwidth}
%     		\centering
%     		\raisebox{-.5\height}{\includegraphics[width=\linewidth]{figs/uncer-likelihood.png}} 
%     \end{minipage} & 1 \\
%     \hline \\
%     \begin{minipage}[b]{1.2\columnwidth}
%     		\centering
%     		\raisebox{-.5\height}{\includegraphics[width=\linewidth]{figs/uncer-likelihood.png}} 
%     \end{minipage} & 1 \\
%     \bottomrule
 
% \end{tabular}
% \end{spacing}
% \end{figure*}

\section{Protocols}
\label{appendix:protocol}

\subsection{Definition of Notations}
\label{appendix:notation}

\begin{table*}[!ht]
  \centering
  \small
  % \resizebox{0.9\linewidth}{!}
  {\begin{tabular}{cc}
    \hline
    \textbf{Notation} & \textbf{Description} \\
    \hline
    $\mathcal Q$ & Dataset containing $n$ Question-Answering pairs. ($|\mathcal Q| = n$) \\
    $\mathcal P$ & Set of few-shot exemplars. \\
    % $\mathcal A$ & Set of decoding strategies. \\
    $\boldsymbol x_i$ & The $i$-th question sample in $\mathcal Q$. \\
    $\boldsymbol {\hat y}_i$ & The $i$-th ground-truth answer in $\mathcal Q$. \\
    ${\boldsymbol {y}_i}^{(k)}$ & The $k$-th sampled response to the $i$-th question in $\mathcal Q$. \\
    $\boldsymbol {p}_k$ & $k$-th few-shot exemplar to sample ${\boldsymbol {y}_i}^{(k)}$. \\
    % $\boldsymbol {a}^{(k)}_i$ & Selected decoding strategy to sample $\boldsymbol {y}^{(k)}_i$. \\
    $\boldsymbol {Y}_i$ & Answering set containing $K$ sampled response $\left \{ {\boldsymbol {y}_i}^{(k)} \right \}$ for the $i$-th question $\boldsymbol x_i$. \\
    ${z_i}^{(k)}$ & The label of ${\boldsymbol {y}_i}^{(k)}$ (${z_{i}}^{(k)}\in\left \{ 0,1 \right \}$, 1 for \textit{True} and 0 for \textit{False}). \\
    $\boldsymbol {Z}_i$ & Label set corresponding to $\boldsymbol {Y}_i$. \\
    $c_i$ & The confidence score for the $i$-th question $\boldsymbol x_i$. \\
    $e_i$ & The semantic entropy for the $i$-th question $\boldsymbol x_i$. \\
    $\mathcal D$ & Constructed \textsc{UAlign} training set containing $N$ tuple samples $(\boldsymbol x_i, \boldsymbol Y_i, \boldsymbol Z_i, \boldsymbol {\hat y}_i, c_i, e_i)$. \\
    $\tau$ & Uncertainty estimation model trained to calculate confidence score by feeding $\boldsymbol{x}$. \\
    $\mu$ & Uncertainty estimation model trained to calculate semantic entropy by feeding $\boldsymbol{x}$. \\
    $\theta$ & Binary classifier by feeding $(\boldsymbol{x}, c, e, \boldsymbol y)$ as the reward model. \\
    $\mathcal L_{\mathcal M}$ & Training loss functions for three models respectively where $\mathcal M\in \left \{ \tau, \mu, \theta \right \}$. \\    
    $r$ & Final reward signal consisted of reward score $r_1$ and KL-penalty $r_2$. \\
    $\beta$ & Coefficient for the KL-penalty $r_2$. \\
    $\pi_{\theta}$ & Policy model to be optimized using $r$ by PPO. \\
    $\pi_o$ & Reference model initialized by the original policy. \\
    $T$ & Sampling temperatue. \\
    $K$ & Number of sampled responses. \\
    $N$ & Number of QA pairs. \\
    \hline
  \end{tabular}}
  \caption{Summarized notations in this work.}
\label{table:notation}
\end{table*}

The definitions of the notations in this work are summarized in Table \ref{table:notation}.

\subsection{Terminology Use}

\begin{itemize}
    \item In this work, ``\textsc{UAlign}'' in small caps font specifically indicates the proposed framework, which indicates methodology like \textsc{UAlign} dataset, \textsc{UAlign} SFT and \textsc{UAlign} PPO.
\end{itemize}

\section{Method Specification and Supplement}

\subsection{Model Update during PPO}
\label{appendix:ppo}
During the PPO process, only the policy model $\pi_{\theta}$ is optimized while the uncertainty models $\mu, \tau$ do not need to be updated because the reward model $\theta$ updates are offline.
As discussed and demonstrated in Sec. \ref{ssec:measure} and Table \ref{table:reward}, uncertainty models are directly associated with and benefit the reward model. 
In our UAlign PPO algorithm, by incorporating the two uncertainty estimations, the reward model $\theta$ can provide more precise reward scores, thereby guiding LLMs $\pi$ to generate more factual responses.
Since the reward model is offline during PPO, the uncertainty models also do not require online updates.

In addition, due to the KL-divergence constraint, the knowledge distribution of policy LLMs may not diverge too much from the initial policy.
Both uncertainty models and reward models are trained on data generated by sampling from the vanilla LLMs, and their combined effect is to elicit the LLMs' capacity for factual expression, evolving towards improved reliability. 
During the PPO process, with the KL-divergence constraint in Equation \ref{eq:ppo}, the knowledge distribution of policy LLMs may not shift too much from the initial policy. 
We demonstrate the accuracy-based confidence distribution of Llama-3 before and after \textsc{UAlign} training on TriviaQA validation sets as follows.

\begin{table}[ht]
    \centering
    \small
    {\begin{tabular}{ccc}
    \toprule
    \bf Conf. Range & \bf Before UAlign & \bf After UAlign \\
    \hline
    $[0, 0.25)$ & 2404 & 2116 \\
    $[0.25, 0.5)$ & 1786 & 1628 \\
    $[0.5, 0.75)$ & 1509 & 1747 \\
    $[0.75, 1.0]$ & 4261 & 4469 \\
    \bottomrule
    \end{tabular}}
    \caption{Accuracy-based confidence distribution of Llama-3 before and after \textsc{UAlign} training on TriviaQA validation sets.}
\label{table:conf_dist}
\end{table}

Since the knowledge distribution does not shift too much from the initial policy, we can still achieve good performance without updating the uncertainty model for simplicity. 
Compared to traditional RLHF that solely utilize reward models, our proposed \textsc{UAlign} introduces uncertainty models that leverage knowledge boundary representations to benefit reward model and finally enhance LLMs, leading to signification improvements in reliability and generalization of knowledge QA tasks.

\subsection{Computation Cost of Constructing UAlign Dataset}
\label{appendix:train_cost}

As mentioned in Sec. \ref{ssec:sample_num}, we test the time costs to construct the UAlign dataset $\mathcal D$ in different sampling numbers.
We present different sampling time costs on 10000 QA samples on Llama-3-8B and Llama-3-70B \cite{llama3modelcard} \footnote{\href{https://huggingface.co/meta-llama/Meta-Llama-3-70B}{https://huggingface.co/meta-llama/Meta-Llama-3-70B}} on 4×40G A100 GPUs loaded in fp16.
We have tried to address the computation cost problem by introducing many effective acceleration or quantization packages like vllm \footnote{\href{https://github.com/vllm-project/vllm}{https://github.com/vllm-project/vllm}}, bitsandbytes \footnote{\href{https://github.com/bitsandbytes-foundation/bitsandbytes}{https://github.com/bitsandbytes-foundation/bitsandbytes}}, etc that are widely used to drive test time scaling law \citep{snell2024scalingllmtesttimecompute}.
As presented in Table \ref{table:sample_cost}, the results demonstrate the efficiency of our proposed \textsc{UAlign} method even though scaling on larger models.

\begin{table}[ht]
    \centering
    \small
    {\begin{tabular}{ccc}
    \toprule
    \bf Model & \bf Sampling Number & \bf Time Cost \\
    \hline
    \multirow{4}{*}{Llama-3-8B} & 1 & 18 min \\
     & 4 & 22 min \\
     & 7 & 24min \\
     & 10 & 25min \\
     \hline
    \multirow{4}{*}{Llama-3-70B} & 1 & 1h 18min \\
     & 4 & 1h 33min \\
     & 7 & 1h 40min \\
     & 10 & 2h 12min \\
    \bottomrule
    \end{tabular}}
    \caption{Time cost in different sampling numbers of \textsc{UAlign} on Llama-3-8B and Llama-3-70B.}
\label{table:sample_cost}
\end{table}

In addition, the relatively low computation costs when sampling can be attributed that experiments are conducted on knowledge QA datasets. 
The answer spans are entity-level and each answer only needs to generate a few tokens.
Since the output form is relatively simple, sampling ten times is sufficient and cost-saving to accurately fit the knowledge boundaries as presented in Sec. \ref{ssec:sample_num}.

Furthermore, Test Time Scaling Law \citep{snell2024scalingllmtesttimecompute} has attracted much attention recently which proposed to consider allocating more computation resources in inference to generate high-quality responses.
These LLMs' self-generated data can be further used for LLM training to self-improve LLMs \citep{gulcehre2023reinforcedselftrainingrestlanguage}.
Many works validate that incorporating data multiply sampled on LLMs in inference can benefit LLMs for further improvements, which is a new trend for LLM training. 
In this way, our proposed \textsc{UAlign} provides a novel insight of the test time scaling law to represent knowledge boundaries by calculating uncertainty estimations on the sampled responses, and further explicitly leverages the uncertainty estimations for factuality alignment, heralding a promising view of test time scaling law.
Although few additional computation costs are required, our proposed \textsc{UAlign} is still efficient to be utilized practically with significant reliability improvements.

\subsection{Computation Cost of Inference of UAlign}
\label{appendix:infer_cost}

Following \ref{appendix:train_cost}, we subsequently analyze the computation cost during inference of \textsc{UAlign}.
Our proposed \textsc{UAlign} barely increases additional inference memory and time budget as follows. 

First, uncertainty models also share the base LLMs with their respective plug-in LoRA modules with rank $r$=4. Additional parameters introduced only account for less than 1\% of the base model parameters.

Second, uncertainty models only predict two tokens of uncertainty estimations in inference.
We report the inference time cost on four test sets of vanilla Llama-3 which only generates the answer to the question and \textsc{UAlign} trained Llama-3 which predicts uncertainty estimations and then generates the answers on a single A100 GPU Card. 

\begin{table}[ht]
    \centering
    \small
    {\begin{tabular}{ccc}
    \toprule
    \multirow{2}{*}{\bf Dataset} & \multicolumn{2}{c}{\bf Time Cost} \\ 
    & \bf Vanilla ICL & \bf UAlign \\
    \hline
    \bf TVQA & 58 min & 1h 6min \\
    \bf NQ-Open & 28 min & 32m in \\
    \bf SciQ & 6 min & 7 min \\
    \bf LSQA & 5 min & 6 min\\
    \bottomrule & 
    \end{tabular}}
    \caption{Inference time cost on four test sets of Llama-3 using vanilla ICL prompt-based and \textsc{UAlign} methods.
    Note the inference time costs on all the baseline methods in Sec. \ref{ssec:baseline} are comparable to the vanilla ICL prompt-based baseline method.}
\label{table:infer_cost}
\end{table}
    
Therefore, with the slight increase in additional memory and time cost in inference, \textsc{UAlign} significantly outperforms other baseline methods, demonstrating the reliability and efficiency on such tasks.

% \clearpage

% \section{Cases Analysis}
% \label{appendix:cases}

% \input{appendix/cases}

\section{Dataset Details}
\label{appendix:dataset}

\paragraph{TriviaQA}
The TriviaQA dataset \cite{joshi-etal-2017-triviaqa} \footnote{\href{https://huggingface.co/datasets/mandarjoshi/trivia_qa}{https://huggingface.co/datasets/mandarjoshi/trivia\_qa}} is a comprehensive reading comprehension dataset of QA resource consisting of approximately 650,000 question-answer-evidence triples sourced from 95,000 documents on Wikipedia and various other websites.
This dataset is distinguished by its complexity and serves as an effective benchmark for evaluating machine comprehension and open-domain QA systems. 
Unlike standard QA benchmark datasets, where answers are directly retrievable, TriviaQA presents a more rigorous challenge as it requires deeper inference to derive answers.

When constructing the \textsc{UAlign} dataset, we pre-process and extract 76,523 QA samples from the TriviaQA training set and 9,960 from the development set to contribute to the \textsc{UAlign} training and in-domain test set respectively.
Since approximating the knowledge distribution of a question requires multiple sampling where the computation cost is linearly increasing with the sampling time $K$, to simplify the setup and conserve computation resources, we conducted experiments using half of the training data points from the original dataset.

\paragraph{SciQ}

The SciQ dataset \citep{welbl2017sciq} \footnote{\href{https://huggingface.co/datasets/allenai/sciq}{https://huggingface.co/datasets/allenai/sciq}} contains 13,679 crowd-sourced science exam questions about physics, chemistry and biology, among others.
The original dataset was divided, with 11,679 samples allocated as the training set and an additional 1,000 samples designated as the validation set.
These were subsequently incorporated into our \textsc{UAlign} training set and in-domain test set, respectively.

\paragraph{NQ-Open}

The NQ-Open dataset is derived from Natural Question \citep{kwiatkowski-etal-2019-natural}
\footnote{\href{https://huggingface.co/datasets/google-research-datasets/nq_open}{https://huggingface.co/datasets/google-research-datasets/nq\_open}}, which is a QA dataset consisting of real queries issued to the Google search engine.
We employ the training and development set of NQ-Open, which contains 87,925 and 3,610 samples respectively, to further enhance the \textsc{UAlign} training and in-doamin test set.
Since data construction is highly expensive, we also randomly sample half of the QA pairs from the source training data. 
We mix the selected training samples to construct the \textsc{UAlign} dataset, which is further used for U2Align SFT+PPO training.

\paragraph{LSQA}

The LSQA dataset is a multilingual knowledge-intensive QA dataset pertaining to language-dominant knowledge covering specific social, geographical, and cultural language contexts for the UK \& US, France, China, Japan, and Thailand respectively.
In this study, we only input the QA pairs in English from each LSQA subset which includes 1,025 samples as the out-of-domain test set.

\section{Evaluation Details}
\label{appendix:evaluate}

\subsection{Precision and Truthfulness}
\label{appendix:prec_truth}

\paragraph{Explanations and Equations}

\begin{table}[ht]
    \centering
    \small
    {\begin{tabular}{ccc}
    \toprule
    \bf Notation & \bf Indication \\
    \hline
    KC & Known and answered correctly \\
    KI & Known but answered incorrectly \\
    KR & Known but refused to answer \\
    UC & Unknown but answered correctly \\
    UI & Unknown but answered correctly \\
    UR & Unknown and refused to answer \\
    \bottomrule
    \end{tabular}}
    \caption{Denotation of different answer types.}
\label{table:prec_truth}
\end{table}

As defined in Table \ref{table:prec_truth}, "\textit{Truthfulness}" is the proportion of questions the LLM either the known answered correctly or the unknown refused to answer, which measures the honesty of LLMs. 
Some unknown but correctly guessed answers will not be included.
The equation of \textit{Truthfulness} is as follows.

% \begin{footnotesize}
\begin{align}
    \mathrm{Truth}&\mathrm{fulness} = \nonumber\\
    &\mathrm{\frac{UR+KC}{KC+KI+KR+UC+UI+UR}}
\end{align}
% \end{footnotesize}

\textit{Precision}  is defined as the proportion of correctly answered questions among all the known questions, representing LLMs’ ability to accurately express their known factual knowledge.
The equation of \textit{Precision} is as follows.

\begin{align}
    \mathrm{Precision = \frac{KC}{KI+KC+KR}}
\end{align}

\paragraph{Clarifications of Use of Truthfulness}
To avoid the over-conservative problem incurred by using precision only, we employ "truthfulness" as complementary to measure the proportion of questions the model either known answered correctly or unknown refused to answer, which reflects the honesty of the model.
Therefore, as demonstrated in Sec. \ref{sec:result} and Table \ref{table:main_exp}, the previous methods like R-Tuning which may lead models to be overly conservative perform well in precision but poor in truthfulness.
The employed two metrics of precision and truthfulness can comprehensively measure the reliability of different methods, thereby comprehensively demonstrating the superiority of our method over other baselines from these two perspectives.

\subsection{Accuracy}
\label{appendix:accu}

For closed-book QA evaluation, we observe that simply applying EM may misjudge the correct answers.
We compare several variants of EM as in Table \ref{table:em} and report their successful judgments on responses of 20 selected samples that are misjudged using EM, where PEM, RRM, and PREM indicate Positive-EM, Recall-EM, and Positive-Recall-EM and the mathematical explanations are presented in Table \ref{table:em}.
Upon human discrimination, EMPR exhibits the lowest failure rate and is therefore selected as the evaluation metric for this work.

\begin{table}[ht]
    \centering
    \small
    {\begin{tabular}{ccc}
    \toprule
    \bf Variant & \bf Explanation & \bf \# Fail \\
    \hline
    EM & $\boldsymbol y \equiv \boldsymbol {\hat y}$ & 20 \\
    PEM & $\boldsymbol y\in \boldsymbol {\hat y}$ & 16 \\
    REM & $\boldsymbol {\hat y} \in \boldsymbol y$. & 6 \\
    PREM & $\boldsymbol y\in \boldsymbol {\hat y} \vee \boldsymbol {\hat y} \in \boldsymbol y$. & 2 \\
    \bottomrule
    \end{tabular}}
    \caption{Number of failed judgments by human check for different EM variants.}
\label{table:em}
\end{table}

\begin{figure*}[t!]
  \centering
  \includegraphics[width=0.88\textwidth]{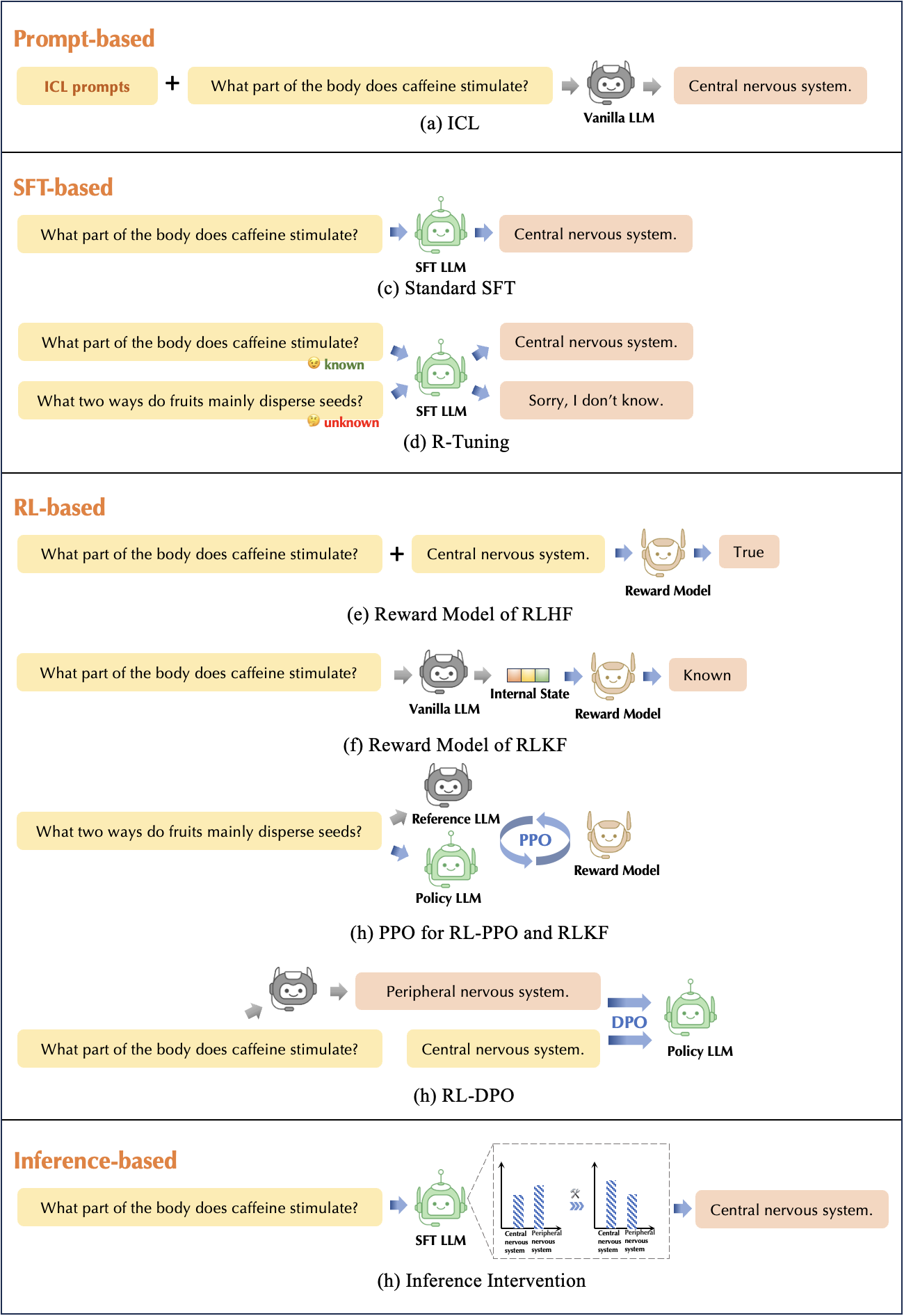}
    \caption{Illustration of several baselines as in Sec. \ref{ssec:baseline}.}
  \label{fig:baseline}
\end{figure*}

\subsection{AUROC}
\label{appendix:auroc}
Area Under the Receiver Operator Characteristic Curve (AUROC) assesses the effectiveness of confidence estimation \cite{filos2019benchmarking} by quantifying how likely a randomly chosen correct answer possesses a higher confidence score than an incorrect one, yielding a score within the range of [0, 1], implemented by \texttt{sklearn} toolkit \footnote{\href{https://github.com/scikit-learn/scikit-learn/blob/main/sklearn/metrics/_ranking.py}{https://github.com/scikit-learn/scikit-learn/blob/main/sklearn/metrics/\_ranking.py}}.
A higher AUROC score implying higher reliability is preferred.

\section{Baseline Details}
\label{appendix:baseline}

\paragraph{Prompt-based}
For all in-context learning methods, we extract the examples from the respective training set to mitigate the knowledge distribution shift between different datasets.
For example, the demonstrated examples in Appendix \ref{appendix:prompt_example} are derived from the TriviaQA training set and are specifically used when inferring on the TriviaQA validation set.
For LSQA without the training set, we use the same examples as TriviaQA as their knowledge domains largely overlap.

\begin{itemize}
    \item \textbf{ICL}: Few-shot prompts containing $m$ examples are utilized for answer generation with temperature $T=0.2$ where $m$ is set to 2 as presented in the Template \ref{temp:icl}.
    \item \textbf{ICL-IDK}: Two examples are included in the few-shot prompt while one is selected from the ICL-used example, and another is an unknown question whose answer is revised to ``\textit{Sorry, I don't know.}'' as presented in the Template \ref{temp:icl-idk}.
    \item \textbf{ICL-CoT}: We also employ the Chain-of-Thought in few-shot examples by recalling the relevant knowledge piece of LLMs and incorporating it into thinking steps before answering the question as presented in the Template \ref{temp:icl-cot}.
    \item \textbf{SFT}: The standard supervised fine-tuning (SFT) is implemented by minimizing the negative log-likelihood of the ground-truth $\boldsymbol {\hat y}$ conditioned on input question $\boldsymbol{x}$ on model $\pi$.

    \begin{align}
        \label{eq:sft}
        \arg \min_{\pi}\mathcal L_{\mathrm{SFT}} = -\mathbb{E}_{(\boldsymbol{x}_i,\boldsymbol {\hat y}_i)\sim\mathcal D}\left [ \log p_{\pi}(\boldsymbol {\hat y}|\boldsymbol x) \right ]
    \end{align}
    \item \textbf{R-Tuning}: R-Tuning \citep{zhang2024rtuning} is implemented in the same way as SFT which only revises the ground-truth label of unknown questions to the refusal answers.
    The unknown questions are determined if all the sampled responses in the \textsc{UAlign} dataset are incorrect.
    \item \textbf{RL-PPO}: Following \citep{ouyang2022training}, we develop the RL-PPO by training a reward model using the LLM-generated incorrect responses as negative samples.
    Then we conduct the PPO \citep{schulman2017proximal} algorithm with the obtained reward model.
    In other word, the RL-PPO baseline is a variant of \textsc{UAlign} which discards the uncertainty estimations.
    \item \textbf{RLKF}: Following \citep{liang-etal-2024-learning}, we employ the RLKF baseline by training the reward model on the LLMs' internal states with the knowledge probes and further conduct PPO using the reward model.
    The knowledge probing setting and implementations are referred to as \citet{liang-etal-2024-learning}.
    \item \textbf{RL-DPO}: All \citet{tian2024finetuning,lin2024flame,zhang2024selfalignment} focus on long-context generation like biography.
    We still utilize the LLMs' generated incorrect responses as negative samples to construct the preference data to conduct the DPO \citep{rafailov2023direct} algorithm.
    \item \textbf{ITI}: We replicate \citep{li2023inferencetime} by training a head probe in the attention layer to intervene in the activations to the ``truthfulness'' direction.
    To be consistent with the original work, we also train the head on TruthfulQA \citep{lin-etal-2022-truthfulqa} with our prepared \textsc{UAlign} dataset to decode in the ``truthfulness'' direction.
    Then we further train the LLM using LoRA by SFT to adapt QA tasks.
    Therefore, the replicated ITI can be regarded as conducting SFT on LLMs with an additional ``truthfulness'' head.
\end{itemize}

\section{Prompt Template}
\label{appendix:prompt}

% \begin{table}[h]
%   \footnotesize
%   \centering
%   \begin{tabular}{
%   >{\columncolor{almond}}p{6.8cm}
%   }
%     \multicolumn{1}{c}{\textbf{{Vanilla Question-Answering}}} \\
%     You are an excellent Question-Answering assistant. Please answer the following question based on your knowledge.\\
%     \#\#\# Question \#\#\#: {\texttt{\{input\_question\}}}\\
%     \#\#\# Answer \#\#\#: \\
%  \\
%   \end{tabular}
%   \caption{A table with a pink background}
%   \label{tab:pink_background}
% \end{table}

\begin{quote}
    \begin{tcolorbox}[colback=almond!20!white, colframe=almond!60!black, title=\textbf{ICL Prompt}]
    \label{temp:icl}
        \small
        You are an excellent Question-Answering assistant. Please answer the following question based on your knowledge.\\
        \\
        \#\#\# Question \#\#\#: {\texttt{\{demo\_question\_1\}}}\\
        \#\#\# Answer \#\#\#: {\texttt{\{demo\_answer\_1\}}}\\
        \\
        \#\#\# Question \#\#\#: {\texttt{\{demo\_question\_2\}}}\\
        \#\#\# Answer \#\#\#: {\texttt{\{demo\_answer\_2\}}}\\
        \\
        \#\#\# Question \#\#\#: {\texttt{\{input\_question\}}}\\
        \#\#\# Answer \#\#\#: \\
    \end{tcolorbox}
    
    \begin{tcolorbox}[colback=almond!20!white, colframe=almond!60!black, title=\textbf{ICL-IDK Prompt}]
    \label{temp:icl-idk}
        \small
        You are an excellent Question-Answering assistant. Please answer the following question based on your knowledge.\\
        \\
        \#\#\# Question \#\#\#: {\texttt{\{demo\_question\_1\}}}\\
        \#\#\# Answer \#\#\#: {\texttt{\{demo\_answer\_1\}}}\\
        \\
        \#\#\# Question \#\#\#: {\texttt{\{demo\_question\_2\}}}\\
        \#\#\# Answer \#\#\#: {\texttt{\{refusal\}}}\\
        \\
        \#\#\# Question \#\#\#: {\texttt{\{input\_question\}}}\\
        \#\#\# Answer \#\#\#: \\
    \end{tcolorbox}
    
    \begin{tcolorbox}[colback=almond!20!white, colframe=almond!60!black, title=\textbf{ICL-CoT Prompt}]
    \label{temp:icl-cot}
        \small
        You are an excellent Question-Answering assistant. Please answer the following question based on your knowledge.\\
        \\
        \#\#\# Question \#\#\#: {\texttt{\{demo\_question\_1\}}}\\
        \#\#\# Recall \#\#\#: {\texttt{\{knowledge\_1\}}}\\
        \#\#\# Answer \#\#\#: {\texttt{\{demo\_answer\_1\}}}\\
        \\
        \#\#\# Question \#\#\#: {\texttt{\{demo\_question\_2\}}}\\
        \#\#\# Recall \#\#\#: {\texttt{\{knowledge\_2\}}}\\
        \#\#\# Answer \#\#\#: {\texttt{\{demo\_answer\_2\}}}\\
        \\
        \#\#\# Question \#\#\#: {\texttt{\{input\_question\}}}\\
        \#\#\# Answer \#\#\#: \\
    \end{tcolorbox}
\end{quote}

\section{Training Setting Details}
\label{appendix:training}

To conserve memory overhead and accelerate computation, all the models are quantified using \texttt{float16 (fp16)} to load and save parameters during both the training and inference phases.
During the training stage, the batch sizes for the LLM, uncertainty estimation models, and reward models are set at 4, 16, and 16, respectively. 
The initial learning rate of 1e-4 is utilized with the 0.05 warm-up ratio and 0.01 weight decay of the ADAM optimizer.
We set the training epoch to 2 and ensure that all the models can be trained to convergence by increasing additional training steps if necessary.
The dropout rate is set at 0.05 during all model updates to reduce overfitting.
In the RL phase, all the hyper-parameters related to PPO algorithm are default values by the \texttt{trl} PPOConfig recipe \footnote{\href{https://github.com/huggingface/trl/blob/main/trl/trainer/ppo_config.py}{https://github.com/huggingface/trl/blob/main/trl/trainer/\\ppo\_config.py}} except the epoch, learning rate, and batch size which are set at 2, 1e-5, and 2, respectively.

\section{Detailed Related Works}
\label{appendix:relate}

\subsection{Knowledge Boundary}

Previous works investigate the knowledge boundary to identify the known level of a knowledge piece of LLMs by quantifying the confidence or uncertainty estimations like output consistency \citep{cheng2024aiassistants}, prompting methods \citep{ren2023investigating}, or knowledge probing \citep{ji-etal-2024-llm}.
Researchers are examining the limits of parametric knowledge in LLMs with the objective of delineating the extent of the LLMs' knowledge and identifying their capability boundaries.
Present studies on the knowledge boundary primarily focus on measuring the knowledge boundaries using confidence or uncertainty estimations on specialized tasks.
The ambiguity of knowledge boundaries can be attributed to the knowledge distribution learned from the pre-training stage or the influence of external knowledge leading to knowledge conflict \citep{xu-etal-2024-knowledge-conflicts} and inconsistency \citep{xue-etal-2023-improving}.

\begin{figure*}[!ht]
  \centering
  \includegraphics[width=0.95\textwidth]{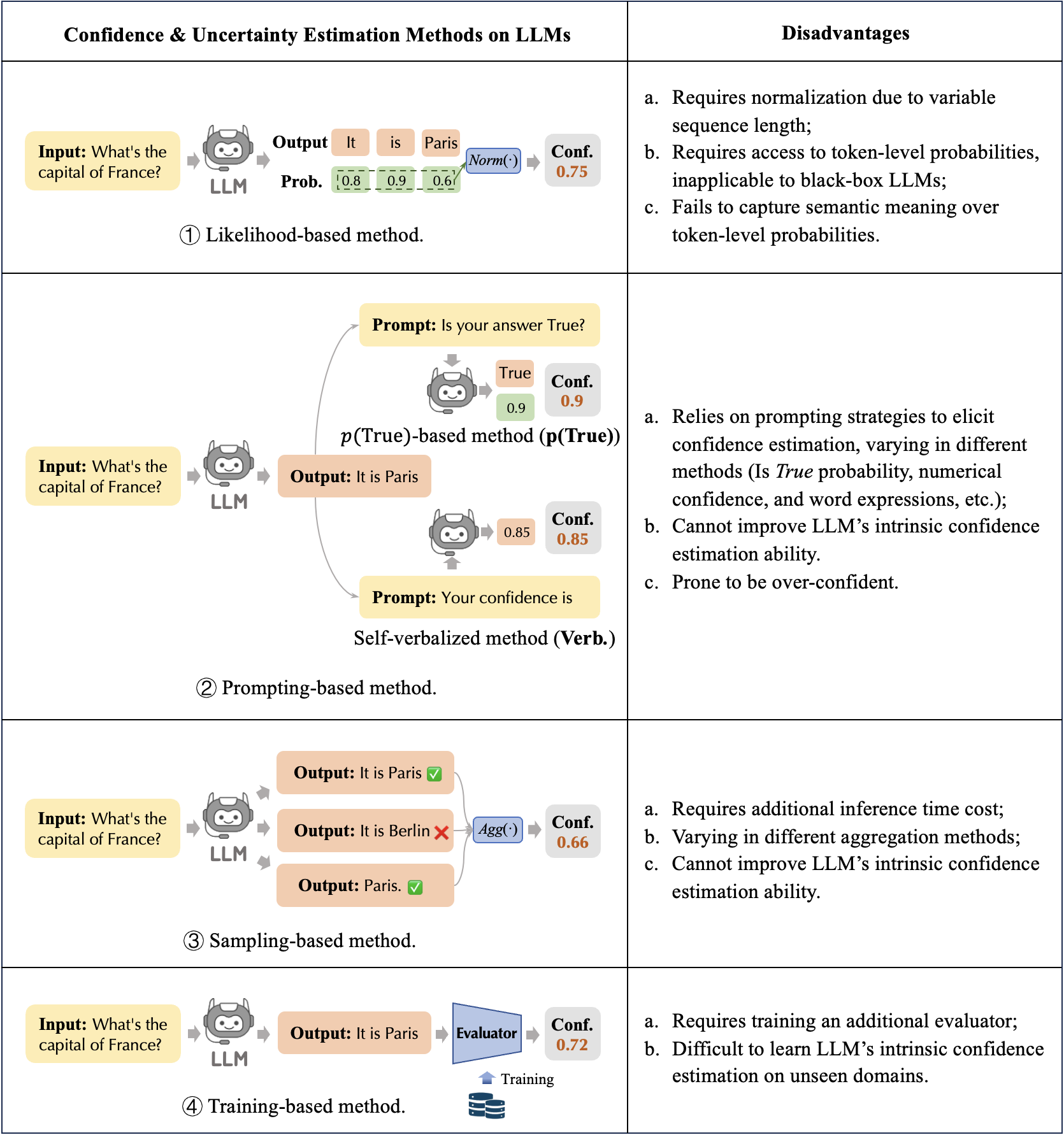}
    \caption{Several uncertainty estimation methods for Generative LLMs.}
  \label{fig:uncer}
\end{figure*}

\subsection{Uncertainty Estimation of LLMs}

To alleviate over-confidence and enhance the reliability of LLMs, reliable uncertainty estimation is essential to determine whether a question is known or not to the LLM \citep{geng2023survey}.
Both \textit{Uncertainty} and \textit{Confidence} estimations can indicate the reliability degree of the responses generated by LLMs, and are generally used interchangeably \cite{xiao-etal-2022-uncertainty,chen2023quantifying,geng2023survey,lu2024psv}.
Uncertainty detection is essential for hallucination mitigation on knowledge-based tasks \citep{xiong2024can,varshney2023stitch,she2022cop,wang2024enhancinglargelanguagemodels,vazhentsev-etal-2023-efficient,wang2024rolepromptingguideddomain,manakul-etal-2023-selfcheckgpt}
In this part, we investigate several commonly used \textit{confidence} \& \textit{uncertainty} estimation methods for generative LLMs as mentioned in Sec. \ref{sec:related}.
Specifically, we denote $\mathsf{Conf}(\boldsymbol x, \boldsymbol y)$ as the confidence score associated with the output sequence $\boldsymbol y=[y_1, y_2, \dots, y_N]$ given the input context $\boldsymbol x=[x_1, x_2, \dots, x_M]$.
We also illustrate the summarized estimation methods as well as their disadvantages in Fig. \ref{fig:uncer}.

\paragraph{Likelihood-based Methods:}

Following model calibration on classification tasks \citep{guo2017calibration},
\citet{vazhentsev-etal-2023-efficient,xue2024comprehensive,varshney2023stitch,wang2024selfdc} intermediately quantify sentence uncertainty over token probabilities.
In traditional discriminative models, except likelihood-based methods, confidence estimations also include ensemble-based and Bayesian methods \citep{balaji2017advances,pmlr-v48-gal16,xue2022bayesian,wang2020bayesian,gal2016uncertainty,abdar2021review}, and density-based methods \citep{NEURIPS2018_abdeb6f5}.
However, this likelihood-based method requires access to token probabilities and thus being limited to white-box LLMs.
The likelihood-based confidence is estimated by calculating the joint token-level probabilities over $\boldsymbol y$ conditioned on $\boldsymbol x$.
As longer sequences are supposed to have lower joint likelihood probabilities that shrink exponentially with length, the product of conditional token probabilities of the output should be normalized by calculating the geometric mean by the sequence length \cite{murray-chiang-2018-correcting,malinin2021uncertainty}, and the confidence score can be represented as:

\begin{align}
\label{eq:conf_normal}
    \mathsf{Conf}(\boldsymbol x, \boldsymbol y)=\left ( {\prod^N_ip(y_i|\boldsymbol y_{<i},\boldsymbol x)} \right )^{\frac{1}{N}}
\end{align}

\vspace{-0.5em}

Similarly, the arithmetical average of the token probabilities is adopted in \citet{varshney2023stitch}:

\vspace{-0.5em}
\begin{align}
\label{eq:conf_avg}
    \mathsf{Conf}(\boldsymbol x, \boldsymbol y)=\frac{1}{N}{\sum^N_ip(y_i|\boldsymbol y_{<i},\boldsymbol x)} 
\end{align}

\vspace{-0.5em}

Furthermore, a low probability associated with even one generated token may provide more informative evidence of uncertainty \cite{varshney2023stitch}.
Hence, the minimum of token probabilities is also employed.

\begin{align}
\label{eq:conf_min}
    \mathsf{Conf}(\boldsymbol x, \boldsymbol y) = {\min \left \{ p(y_1|\boldsymbol x), \dots, p(y_N|\boldsymbol y_{<N}, \boldsymbol x) \right \}}
\end{align}
\vspace{-0.5em}

\paragraph{Prompting-based Methods:}
Recently, LLMs' remarkable instruction-following ability \citep{brown2020language} provides a view of instructing LLMs to self-estimate their confidence level to previous inputs and outputs including expressing uncertainty in words \citep{lin2022teaching,zhou-etal-2023-navigating,tian-etal-2023-just,xiong2024can}, or instructing the LLM to self-evaluate its correctness on $p(\mathrm{True})$ \citep{kadavath2022language}.
The $P(\mathrm{True})$ confidence score is implemented by simply asking the model itself if its first proposed answer $\boldsymbol y$ to the question $\boldsymbol x$ is true \cite{kadavath2022language}, and then obtaining the probability $p(\mathrm{True})$ assigned by the model, which can implicitly reflect self-reflected certainty as follows.

\vspace{-0.5em}
\begin{align}
\label{eq:conf_ptrue}
    \mathsf{Conf}(\boldsymbol x, \boldsymbol y)=p(\mathrm{True})=p(\boldsymbol y\ \mathrm{is}\ \mathrm{True}?|\boldsymbol x)
\end{align}

Another method is to prompt LLMs to linguistically express tokens of confidence scores in verbalized numbers or words \cite{lin2022teaching,mielke-etal-2022-reducing,zhou-etal-2023-navigating,tian2023just,xiong2024can}.

The sampling-based method refers to randomly sampling multiple responses given a fixed input $\boldsymbol x$ using beam search or temperature sampling strategies \citep{manakul-etal-2023-selfcheckgpt,xiong2024can,lyu2024calibrating}.
Various aggregation methods are adopted on sampled responses to calculate the consistency level as the confidence score.
Moreover, some uncertainty quantification methods are used to calculate the entropy indicating the dispersion level of multiple outputs \citep{kuhn2023semantic,lin2023generating,nikitin2024kernel}.

\paragraph{Training-based Methods:}
For training methods, an external evaluator trained on specific datasets is introduced to output a confidence score given an input and an output.
The evaluator can be a pre-trained NLI model \citep{mielke-etal-2022-reducing}, or a value head connected to the LLM output layer \citep{lin2022teaching,kadavath2022language}, or the LLM itself \citep{han2024enhancing}.

\paragraph{}
However, both self-verbalized and sampling methods for uncertainty estimations using extrinsic prompting or aggregation strategies with additional time costs fail to improve LLMs' intrinsic capability of uncertainty estimation.
Recent works investigate confidence learning methods to enhance the reliability of LLMs \citep{han2024enhancing}.
\citet{li2023inferencetime} introduces Inference-Time Intervention (ITI) to enhance the truthfulness of LLMs by shifting model activations during inference.
\citet{yang2023improving} proposes an uncertainty-aware in-context learning method leveraging uncertainty information to refine the responses but cannot improve uncertainty estimation.
\cite{zhang2024rtuning} proposes R-tuning to instruct LLMs to refuse unknown questions considering uncertainty estimations as binary indicators.
In contrast, our proposed \textsc{UAlign} framework not only obtains more reliable uncertainty estimations regarding knowledge boundary information but also elicits accurate responses of LLMs.

\subsection{Factuality Alignment of LLMs}

Alignment is a standard procedure to improve LLMs' helpfulness and factuality \citep{bai2022training}.
The main goal of LLM alignment is to guide human preference through Supervised Fine-Tuning (SFT), Reinforcement Learning from Human Feedback (RLHF) \citep{ouyang2022training,bai2022training} or AI feedback \citep{bai2022constitutional},
which may also guide LLMs to output detailed and lengthy responses \citep{singhal2023long} but inevitably encourage hallucination.
Therefore, many works explore to apply RL to improve LLMs’ factuality through Proximal Policy Optimization (PPO) \citep{schulman2017proximal} with the trained reward model \citep{liang-etal-2024-learning,xu2024rejection} or Direct Preference Optimization (DPO) \citet{rafailov2023direct} with the constructed preference dataset \citep{zhang2024selfalignment,lin2024flame} to align with factuality preferences annotated by human beings.
\citet{xu2024rejection} encourages LLM to reject unknown questions using the constructed preference data by leveraging knowledge boundary feedback.
Some works also explore alignment method for LLM safety \cite{yu-etal-2024-cosafe,yu-etal-2024-repalm,yu-etal-2024-popalm}.

\section{Experiments}
\label{appendix:exp}

\subsection{Experiments of Reliability of Uncertainty Estimations}
\begin{figure}[!ht]
  \centering
  \includegraphics[width=0.36\textwidth]{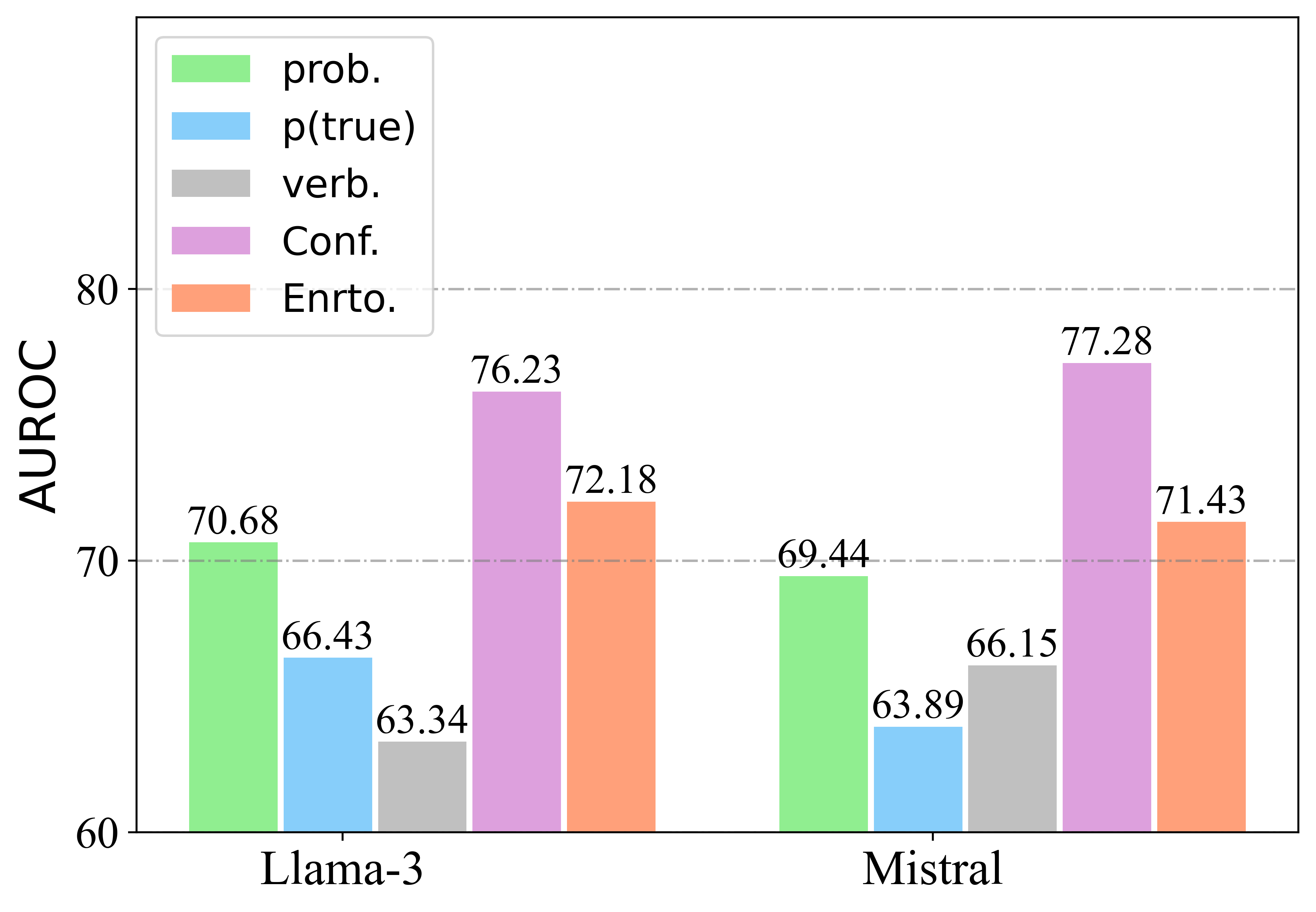}
    \caption{Results of AUORC$\uparrow$ across several confidence/uncertainty estimation methods on SciQ on Llama-3 and Mistral.}
  \label{fig:auroc_sciq}
\end{figure}

\begin{figure}[!ht]
  \centering
  \includegraphics[width=0.36\textwidth]{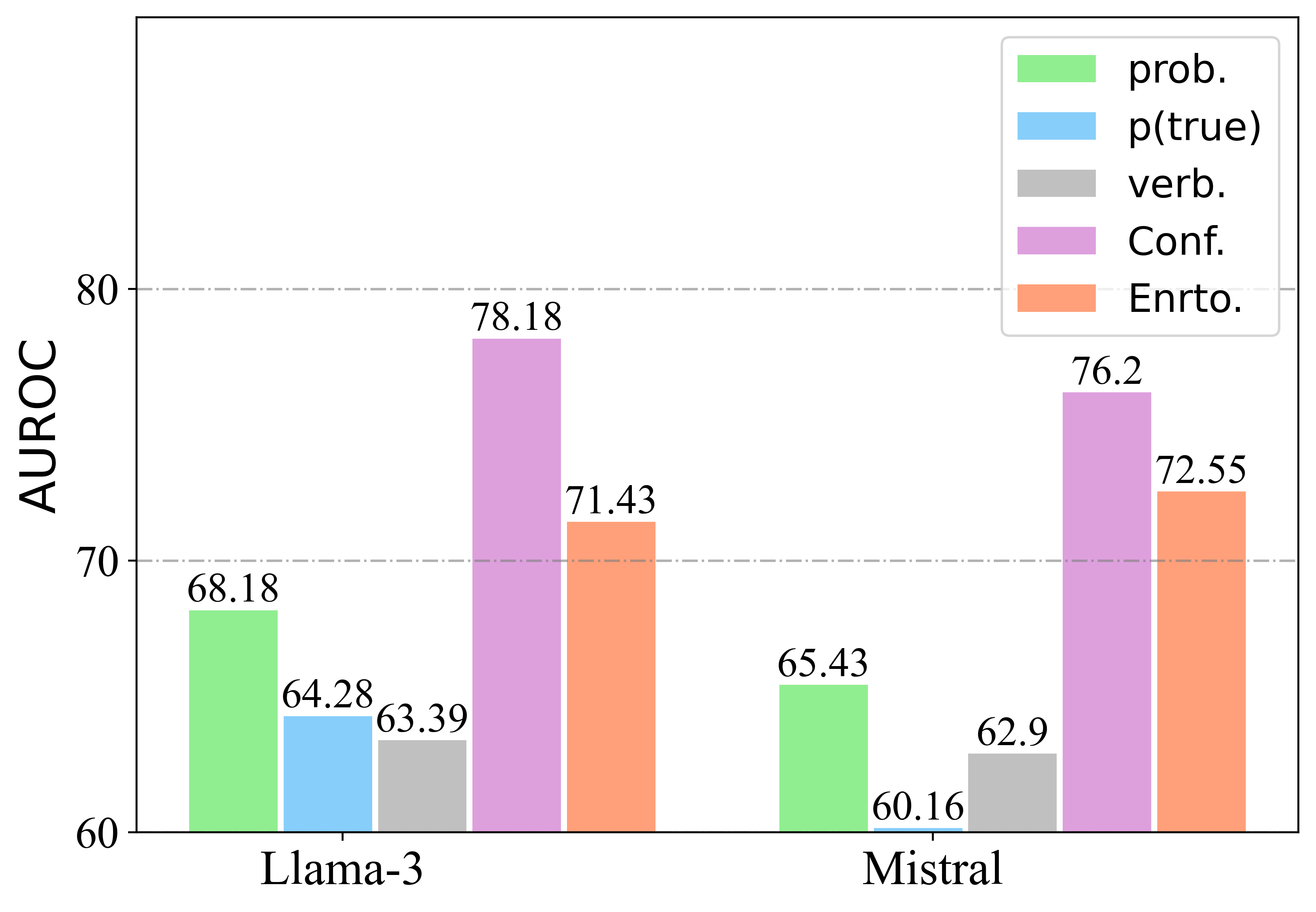}
    \caption{Results of AUORC$\uparrow$ across several confidence/uncertainty estimation methods on NQ-Open on Llama-3 and Mistral.}
  \label{fig:auroc_nqopen}
\end{figure}

\begin{figure}[!ht]
  \centering
  \includegraphics[width=0.36\textwidth]{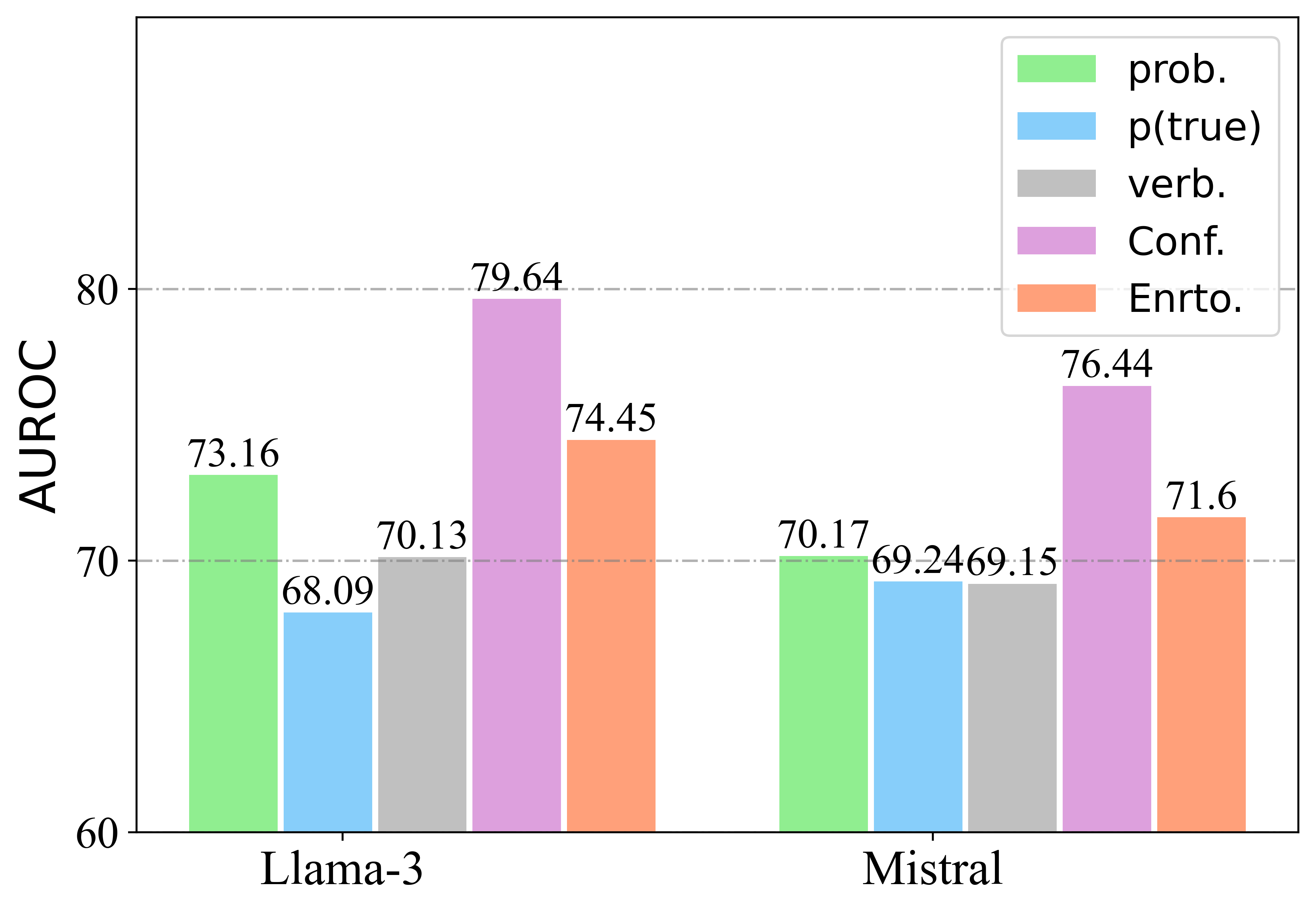}
    \caption{Results of AUORC$\uparrow$ across several confidence/uncertainty estimation methods on LSQA on Llama-3 and Mistral.}
  \label{fig:auroc_lsqa}
\end{figure}

Due to the page limitation in the main part, we present the AUROC performance results of the used confidence and entropy compared with other baseline uncertainty estimations on SciQ, NQ-Open, and LSQA as in Fig. \ref{fig:auroc_sciq}, \ref{fig:auroc_nqopen}, and \ref{fig:auroc_lsqa}.

\section{Few-shot Prompt Examples}
\label{appendix:prompt_example}

10 different few-shot prompts for sampling on TriviaQA are demonstrated in Table \ref{table:example}.

\begin{table*}[!ht]
  \centering
  % \resizebox{0.9\linewidth}{!}
  {\begin{tabular}{cp{13cm}}
    \hline
    \textbf{Examplar ID} & \textbf{Examples} \\
    % \hline
    % \rowcolor{platinum}
    % \multicolumn{2}{c}{\textbf{TriviaQA} \\
    \hline
    1 & Q: Which William wrote the novel Lord Of The Flies? A: Golding. \\
    2 & Q: Where in England was Dame Judi Dench born? A: York, UK. \\
    3 & Q: Neil Armstrong was a pilot in which war? A: Korean. \\
    4 & Q: How many home runs did baseball great Ty Cobb hit in the three world series in which he played? A: None. \\
    5 & Q: Who had a big 60s No 1 with Tossin' and Turnin'? A: Bobby Lewis. \\
    6 & Q: Which Disney film had the theme tune A Whole New World? A: 'Ala' ad Din. \\
    7 & Q: In basketball where do the Celtics come from? A: City of Boston. \\
    8 & Q: Which element along with polonium did the Curies discover? A: Radium. \\
    9 & Q: Who was the Egyptian king whose tomb an treasures were discovered in the Valley of the Kings in 1922? A: Tutanhamon. \\
    10 & Q: Where were the 2004 Summer Olympic Games held? A: Atina, Greece. \\
    \hline
  \end{tabular}}
  \caption{Demonstrations of 1-shot examples for TriviaQA sampling to construct \textsc{UAlign} dataset.}
\label{table:example}
\end{table*}

\begin{table*}[!ht]
  \centering
  % \resizebox{0.9\linewidth}{!}
  {\begin{tabular}{cp{13cm}}
    \hline
    \textbf{Examplar ID} & \textbf{Examples} \\
    % \hline
    % \rowcolor{platinum}
    % \multicolumn{2}{c}{\textbf{TriviaQA} \\
    \hline
    1 & Q: What type of organism is commonly used in preparation of foods such as cheese and yogurt? A: mesophilic organisms. \\
    2 & Q: What phenomenon makes global winds blow northeast to southwest or the reverse in the northern hemisphere and northwest to southeast or the reverse in the southern hemisphere? A: coriolis effect. \\
    3 & Q: Changes from a less-ordered state to a more-ordered state (such as a liquid to a solid) are always what? A: exothermic. \\
    4 & Q: What is the least dangerous radioactive decay? A: alpha decay. \\
    5 & Q: Kilauea in hawaii is the world’s most continuously active volcano. very active volcanoes characteristically eject red-hot rocks and lava rather than this? A: smoke and ash. \\
    6 & Q: When a meteoroid reaches earth, what is the remaining object called? A: meteorite. \\
    7 & Q: What kind of a reaction occurs when a substance reacts quickly with oxygen? A: combustion reaction. \\
    8 & Q: Organisms categorized by what species descriptor demonstrate a version of allopatric speciation and have limited regions of overlap with one another, but where they overlap they interbreed successfully? A: ring species. \\
    9 & Q: Alpha emission is a type of what? A: radioactivity. \\
    10 & Q: What is the stored food in a seed called? A: endosperm. \\
    \hline
  \end{tabular}}
  \caption{Demonstrations of 1-shot examples for SciQ sampling to construct \textsc{UAlign} dataset.}
\label{table:example2}
\end{table*}

\end{document}